\documentclass[sigconf, nonacm, screen]{acmart}
\AtBeginDocument{%
  }

\setcopyright{acmlicensed}
\copyrightyear{2018}
\acmYear{2018}
\acmDOI{XXXXXXX.XXXXXXX}
\acmISBN{}

\acmSubmissionID{}

\usepackage{multirow}
\usepackage{subcaption}
\usepackage{xspace}

\newcommand{\eg}{\textit{e.g.}\xspace}
\newcommand{\ie}{\textit{i.e.}\xspace}

\AtEndPreamble{
    \usepackage[capitalize]{cleveref}
    \crefname{section}{Sec.}{Secs.}
    \Crefname{section}{Section}{Sections}
    \crefname{table}{Tab.}{Tabs.}
    \Crefname{table}{Table}{Tables}
}
\begin{document}

\sloppy

\title[{MHSA: A Lightweight Framework for \underline{M}itigating \underline{H}allucinations via \underline{S}teered \underline{A}ttention in LVLMs}]{MHSA: A Lightweight Framework for \\ \underline{M}itigating \underline{H}allucinations via \underline{S}teered \underline{A}ttention in LVLMs}

\author{Wei Ding}
\orcid{0009-0004-5454-7287}
\authornote{Equal contribution.}
\affiliation{%
  \institution{Tsinghua University}
  \city{Beijing}
  \country{China}
}
\email{teresading999@gmail.com}

\author{Yilin Li}
\orcid{0009-0006-7062-8065}
\authornotemark[1]
\authornote{This work was completed during a visiting period at Tsinghua University.}
\affiliation{%
  % \institution{Beijing University of Posts and Telecommunications}
  \institution{Tsinghua University}
  \city{Beijing}
  \country{China}
}
\email{liyilin784@gmail.com}

\author{Yudong Zhang}
\orcid{0009-0009-6049-603X}
\authornotemark[1]
\authornote{Corresponding authors.}
\affiliation{%
  \institution{Tsinghua University, Tencent}
  \city{Beijing}
  \country{China}
}
\email{zhangyd16@mails.tsinghua.edu.cn}

\author{Ruobing Xie}
\orcid{0000-0003-3170-5647}

\affiliation{%
  \institution{Tencent}
  \city{Beijing}
  \country{China}}
\email{xrbsnowing@163.com}

\author{Jiansheng Chen}
\orcid{0000-0002-2040-7938}

\affiliation{%
  \institution{University of Science and Technology Beijing}
  \city{Beijing}
  \country{China}
}
\email{jschen@ustb.edu.cn}

\author{Xingwu Sun}
\orcid{0009-0008-3222-0901}
\affiliation{%
 \institution{University of Macau}
 \city{Macau}
 \country{China}}
\email{sunxingwu01@gmail.com}

\author{Yu Wang}
\orcid{0000-0001-6108-5157}
\authornotemark[3]
\affiliation{%
  \institution{Tsinghua University}
  \city{Beijing}
  \country{China}}
\email{yu-wang@mail.tsinghua.edu.cn}

% \renewcommand{\shortauthors}{Ding et al.}

\renewcommand\footnotetextcopyrightpermission[1]{}
\settopmatter{printacmref=false} %remove ACM reference format

\begin{abstract}
  Large vision-language models (LVLMs) have achieved remarkable performance across diverse multimodal tasks, yet they continue to suffer from hallucinations, generating content that is inconsistent with the visual input. Prior work DHCP (Detecting Hallucinations by Cross-modal Attention Pattern) has explored hallucination detection from the perspective of cross-modal attention, but does not address hallucination mitigation. In this paper, we propose MHSA (Mitigating Hallucinations via Steered Attention), a lightweight framework that mitigates hallucinations by learning to correct cross-modal attention patterns in LVLMs. MHSA trains a simple three-layer MLP generator to produce corrected attention, guided by supervisory signals from the DHCP discriminator and the LVLM itself. During inference, MHSA mitigates both discriminative and generative hallucinations across various datasets and LVLMs by simply replacing the original cross-modal attention with the corrected one, without modifying any LVLM parameters. By extending cross-modal attention mechanisms from hallucination detection to hallucination mitigation, MHSA offers a novel perspective on hallucination research in LVLMs and helps enhance their reliability.
\end{abstract}


% \keywords{Do, Not, Use, This, Code, Put, the, Correct, Terms, for,


\maketitle

\section{Introduction}
\label{sec:introduction}

Large vision-language models (LVLMs) have demonstrated exceptional capabilities across a wide range of multimodal tasks, including visual question answering, image captioning, and visual reasoning~\cite{liu2024visual, bai2025qwen25vl, chen2024internvl, dai2024instructblip}, building upon earlier large-scale vision-language pre-training efforts such as CLIP~\cite{DBLP:conf/icml/RadfordKHRGASAM21}. By integrating visual encoders with large language models (LLMs) through carefully designed projectors, LVLMs can effectively interpret and reason about visual content~\cite{li2023blip2, zhu2023minigpt4, alayrac2022flamingo}. Despite these impressive achievements, a critical challenge persists: the issue of \emph{hallucinations}, where models generate content that is inconsistent with or absent from the input visual information~\cite{li2023evaluating, rohrbach2018chair}, as further highlighted by recent hallucination and MLLM evaluation benchmarks such as M-HalDetect~\cite{DBLP:conf/aaai/GunjalYB24}, HallusionBench~\cite{DBLP:conf/cvpr/GuanLWXLL0CHYM024}, and MME~\cite{DBLP:journals/corr/abs-2306-13394}. Hallucinations in LVLMs manifest in diverse forms, including object hallucinations (incorrectly identifying objects), attribute hallucinations (misidentifying states, numbers, or actions of objects), and relational hallucinations (erroneously describing spatial or contextual relationships between objects)~\cite{wang2023amber, li2023evaluating}. These hallucinations severely undermine the reliability and trustworthiness of LVLMs in practical applications, making hallucination mitigation a pressing research priority.

\begin{figure}[t]
    \centering
    \includegraphics[width=\linewidth]{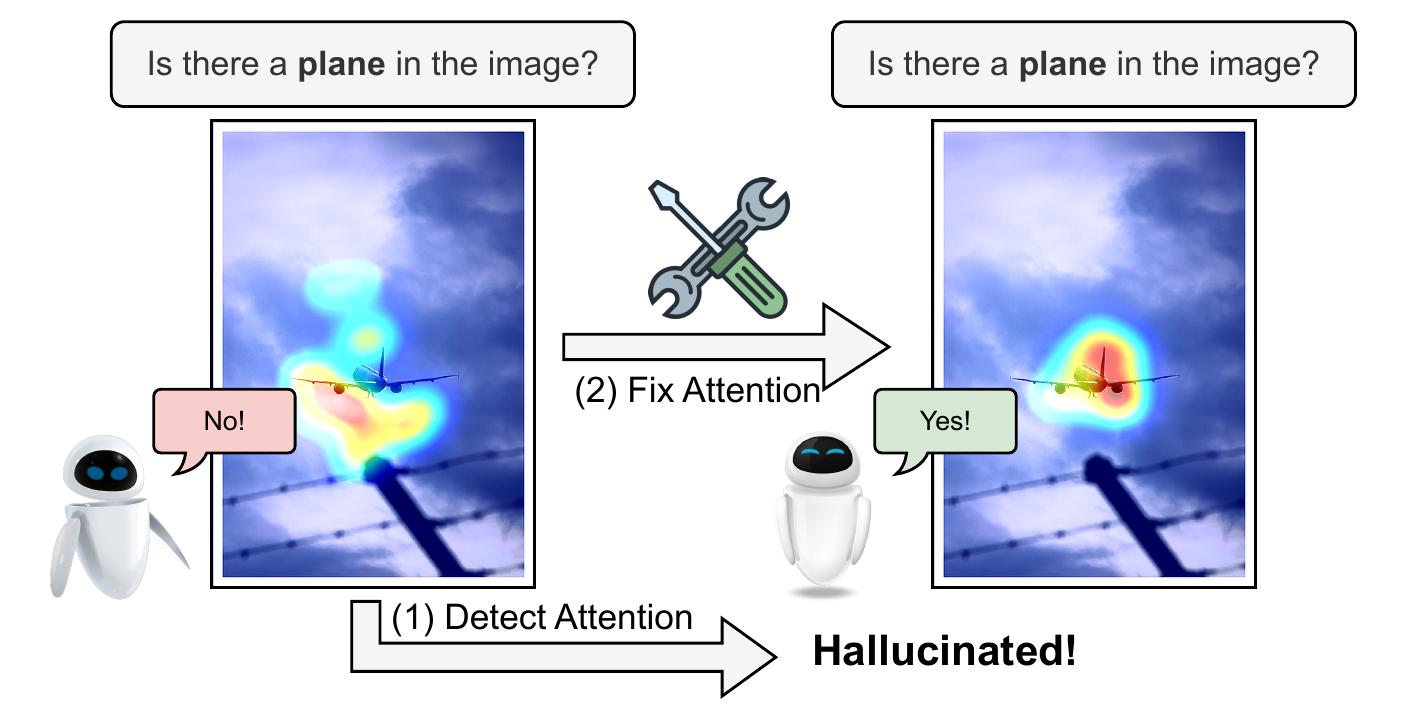}
    \caption{Schematic diagram of the MHSA mechanism. The LVLM first generates hallucinated responses, and the supervision signal from the DHCP discriminator guides the MHSA corrector to repair cross-modal attention, thereby achieving hallucination mitigation.}
    \label{fig:teaser}
    % \vspace{-2em}
\end{figure}

Existing approaches for mitigating hallucinations can be broadly categorized into two classes. \textbf{Training-based methods} involve fine-tuning the LVLM with reinforcement learning from human feedback (RLHF) or instruction tuning to reduce hallucinated outputs~\cite{sun2023aligning, liu2023mitigating, stiennon2020learning}. While effective, these methods require substantial computational resources and high-quality annotated data. \textbf{Inference-time methods} modify the decoding process to suppress hallucinations. However, these methods rely on hand-crafted, fixed heuristic rules (e.g., manually designed penalty functions or static re-weighting schemes) that do not adapt to the specific hallucination signature of each input sample.

Recently, a line of research has explored the relationship between cross-modal attention patterns and hallucinations in LVLMs. Specifically, DHCP~\cite{zhang2025dhcp} reveals that the cross-modal attention weights assigned by generated text tokens to visual tokens exhibit significant and distinguishable differences between hallucinated and non-hallucinated samples. Based on this finding, DHCP trains a lightweight two-layer MLP detector that identifies hallucinations by monitoring cross-modal attention during inference, without any additional LVLM training or extra inference steps. However, DHCP has two key limitations: (1) it can only detect hallucinations but cannot mitigate them; (2) it operates solely at the sentence level and cannot detect hallucinations at the token level. Despite these limitations, the finding opens a promising avenue: if hallucinations can be detected through attention patterns, can they also be \emph{mitigated} by learning to correct these patterns in a data-driven manner?

In this paper, we propose \textbf{MHSA} (\underline{M}itigating \underline{H}allucinations via \underline{S}teered \underline{A}ttention), a lightweight framework that bridges the gap between hallucination detection and mitigation by learning to correct cross-modal attention patterns, as shown in \cref{fig:teaser}. Unlike existing attention manipulation methods (\eg, OPERA~\cite{huang2023opera}, PAI~\cite{liu2024paying}) that apply fixed heuristic rules, MHSA is the first to learn \emph{sample-adaptive} attention corrections through a data-driven training procedure. Specifically, MHSA trains a lightweight MLP-based generator $G$ that takes the original cross-modal attention $\mathbf{A}$ as input and produces a correction term $\Delta\mathbf{A}$, yielding the corrected attention $\mathbf{A}' = \mathbf{A} + \Delta\mathbf{A}$. The pre-trained DHCP detector $D$ is repurposed as a token-level discriminator, providing attention-guidance supervision signals to steer the corrected attention toward non-hallucinatory patterns. The framework is further regularized to preserve the original attention structure and maintain output quality. Importantly, MHSA does not modify any parameters of the LVLM backbone; the lightweight generator and DHCP are trained, and it replaces the cross-modal attention during LVLM inference.

We evaluate MHSA on several mainstream LVLMs, including Qwen2.5-VL, InternVL2-8B, and LLaVA-v1.5. Experimental results demonstrate that MHSA achieves stable and reliable hallucination mitigation on discriminative tasks. By extending DHCP's sentence-level detector to the token level, we further evaluate MHSA on generative tasks, where the results show that MHSA also has the potential to mitigate hallucinations in open-ended generation.

Our contributions can be summarized as follows: (1) We propose MHSA, the first framework that mitigates hallucinations in LVLMs by \emph{learning} sample-adaptive corrections of cross-modal attention patterns. MHSA trains a lightweight three-layer MLP generator with a three-component training objective to produce data-driven corrections guided by a pre-trained hallucination detector, enabling effective and stable hallucination mitigation. (2) We extend MHSA to token-level attention correction for generative tasks by upgrading the attention-based detection to the token level, enabling MHSA to mitigate hallucinations not only in discriminative VQA tasks but also in open-ended image captioning. (3) Extensive experiments across multiple LVLMs and datasets demonstrate that MHSA consistently improves hallucination mitigation metrics on both discriminative and generative tasks. We further provide attention visualizations to validate the causal role of attention correction.

\section{Related Work}
\label{sec:related_work}

\subsection{Hallucination Evaluation in LVLMs}
\label{sec:rw_eval}

Existing evaluation protocols can be broadly divided into \emph{discriminative} and \emph{generative} paradigms.

\textbf{Discriminative evaluation} frames hallucination assessment as binary classification. POPE~\cite{li2023evaluating, lovenia2023nope} converts object presence into Yes/No questions (\eg, ``Is there a [object] in this image?'') and defines three difficulty levels---Random, Popular, and Adversarial---that progressively test the model's susceptibility to co-occurrence priors. AMBER~\cite{wang2023amber} extends this paradigm beyond objects to encompass attribute and relational hallucinations. These benchmarks are efficient and easy to interpret, making them widely adopted for comparing mitigation methods.

\textbf{Generative evaluation} assesses hallucinations in open-ended outputs. CHAIR~\cite{rohrbach2018chair} is a representative metric for image captioning that measures hallucinated objects at both the sentence level (CHAIR$_s$) and instance level (CHAIR$_i$). By examining whether generated object mentions match ground-truth annotations, CHAIR provides a fine-grained measure of hallucination severity in free-form text.

These two paradigms pose fundamentally different challenges for mitigation methods: discriminative tasks require only a binary decision correction, whereas generative tasks demand token-level intervention within an autoregressive sequence. MHSA is designed to address both paradigms through a unified token-level attention correction framework, enabling real-time mitigation during LVLM inference.

\subsection{Hallucination Mitigation}
\label{sec:rw_mitigation}

Hallucination mitigation methods aim to reduce or eliminate hallucinated content in LVLM outputs. We identify two major research lines, each operating at a different level of the model pipeline.

\textbf{Training-based methods} tackle the problem at the \emph{model weight} level by fine-tuning LVLMs on curated data. RLHF-based approaches~\cite{sun2023aligning, stiennon2020learning} leverage human feedback to align model outputs with factual content, while instruction tuning methods~\cite{liu2023mitigating} construct high-quality training data designed to improve model faithfulness. These methods are effective because they directly adjust the model's internal representations to favor factual outputs. However, they depend on substantial computational resources and large-scale, high-quality annotated data, and the learned improvements are tightly bound to the training distribution---hallucination patterns not represented in the training set may remain unaddressed. Moreover, each new model or domain may require a fresh round of expensive fine-tuning.

\textbf{Inference-time methods} intervene at the \emph{output logit} level or \emph{attention pattern} level during inference, requiring no additional training. VCD~\cite{leng2023vcd} contrasts logits from the original and visually distorted inputs, amplifying tokens that genuinely depend on visual evidence. ICD~\cite{wang2024icd} introduces instruction contrastive decoding, distorting the instruction rather than the image. HALC~\cite{DBLP:conf/icml/ChenZLY0Z24} further explores adaptive focal-contrast decoding to reduce object hallucination. These methods provide training-free mitigation; however, they approximately double inference cost due to multiple forward passes, and their effectiveness relies on the assumption that hallucinated and non-hallucinated tokens respond differently to the chosen distortion---an assumption that may not hold uniformly across all hallucination types. Additionally, OPERA~\cite{huang2023opera} introduces over-trust penalty and retrospection-allocation mechanisms to penalize attention columns with abnormally high aggregation, which correlate with hallucination onset. PAI~\cite{liu2024paying} re-weights attention to allocate more focus to image tokens. However, these methods employ heuristic rules (\eg, manually designed penalty functions) rather than learned corrections, and OPERA in particular requires complex backtracking during generation, introducing additional latency. Crucially, these methods modify attention patterns using fixed rules that do not adapt to the specific hallucination signature of each sample.

\textbf{Summary.} Training-based methods are computationally expensive; contrastive decoding doubles inference cost; and attention manipulation methods rely on hand-crafted heuristics that do not adapt to per-sample hallucination signatures. MHSA bridges this gap by learning a lightweight generator that produces \emph{data-driven, sample-adaptive} attention corrections, guided by a pre-trained hallucination detector, without modifying any LVLM backbone parameters.

\subsection{Attention-Based Mechanisms in Trustworthy AI}
\label{sec:rw_detection}

A separate but closely related research line investigates how cross-modal attention patterns can be leveraged to improve trustworthy AI.

In the field of adversarial attacks and defenses, prior work published at ACM MM (BNI track) has utilized cross-modal attention to achieve adversarial sample detection and defense. Specifically, PIP~\cite{DBLP:conf/mm/ZhangXCS024} extracts cross-modal attention from clean and adversarial samples on unrelated probe questions and trains a lightweight support vector machine to detect adversarial samples; F3~\cite{zhang2025f3} achieves adversarial purification by injecting untrained purification noise into adversarial samples through a cross-modal attention guidance mechanism. The success of PIP and F3 demonstrates the potential of cross-modal attention for improving trustworthy AI.

In the field of hallucination detection, prior work also published at ACM MM utilizes cross-modal attention patterns. Specifically, DHCP~\cite{zhang2025dhcp} extracts the average attention weights from generated text tokens to each visual token across all LLM layers and attention heads from both hallucinated and non-hallucinated samples, and then trains a lightweight two-layer MLP detector on these attention patterns, achieving strong hallucination detection performance across discriminative tasks (\eg, POPE, AMBER) and generative tasks (\eg, COCO-Caption) without requiring additional LVLM training or extra inference steps.

The success of DHCP raises a natural question: if hallucinations can be reliably detected through attention patterns, can they also be \emph{mitigated} by directly correcting these patterns? MHSA is motivated precisely by this gap. We adopt DHCP as the detection backbone and repurpose it as a token-level discriminator that provides adversarial supervision for a learned generator, transforming a detection-only tool into a full mitigation framework, enabling hallucination mitigation in both discriminative and generative tasks---a capability that attention-based detection alone cannot provide.

\section{Method}
\label{sec:method}

In this section, we present MHSA, a lightweight framework for mitigating hallucinations in LVLMs via learned attention correction. We first review the preliminaries of cross-modal attention and hallucination detection (\cref{sec:preliminaries}), then formulate the problem (\cref{sec:formulation}), describe the MHSA framework in detail (\cref{sec:mhsa_framework}), and finally discuss the extension to token-level correction for generative tasks (\cref{sec:token_level}).

\subsection{Preliminaries: Cross-Modal Attention and Hallucination Detection}
\label{sec:preliminaries}

We first review the cross-modal attention in LVLMs as defined by DHCP. Consider an LVLM consisting of a visual encoder $f_V$, a projector $f_P$, and an LLM $f_{\text{LLM}}$. Given an image $x_i$ and a text prompt $x_t$, the visual encoder generates features $f_V(x_i)$, which are then projected to obtain $N$ visual tokens fed into the LLM. Let $\mathbf{A}^{(l,h)}_{q \to n}$ denote the attention weight assigned by output position $q$ to the $n$-th visual token at layer $l$ and head $h$.

For discriminative tasks such as POPE (Yes/No question answering), the model generates a single answer token, and DHCP defines the \emph{cross-modal attention} as the attention weights from the first output token to all visual tokens:
\begin{equation}
    \mathbf{A}^{(l,h)}_n = \mathbf{A}^{(l,h)}_{q_{1} \to n}, \quad l = 1, \ldots, L, \; h = 1, \ldots, H, \; n = 1, \ldots, N,
    \label{eq:dhcp_dis}
\end{equation}
yielding a tensor of shape $(L, H, N)$, where $L$ is the number of LLM layers and $H$ is the number of attention heads. For example, LLaVA-v1.5-7B has $(L=32, H=32, N=576)$. For models with dynamic visual token counts such as Qwen2.5-VL-7B, DHCP standardizes the input resolution by resizing all images to $336 \times 336$ pixels before processing, which yields a fixed number of visual tokens ($N=144$) across all samples. This ensures a consistent attention tensor shape for the MLP-based generator and detector.

For generative tasks, DHCP averages the cross-modal attention across all output tokens:
\begin{equation}
    \mathbf{A}^{(l,h)}_n = \frac{1}{M}\sum_{i=1}^{M}\mathbf{A}^{(l,h)}_{q_{i} \to n}, \quad l = 1, \ldots, L, \; h = 1, \ldots, H, \; n = 1, \ldots, N,
    \label{eq:dhcp_gen}
\end{equation}
This means that DHCP can only detect hallucinations at the sentence level and cannot identify which specific token is responsible for the hallucination. Furthermore, the cross-modal attention of hallucinating tokens may be diluted by averaging over the entire sentence, thereby reducing the effectiveness of hallucination detection.

In our framework, we adopt DHCP as the hallucination detector $D$, which operates on cross-modal attention patterns. The detector is a two-layer MLP with a hidden dimension of 128 and an output dimension of 2, representing hallucinatory and non-hallucinatory classes. Given the flattened cross-modal attention as input, $D$ outputs the probability of the sample being hallucinatory or non-hallucinatory:
\begin{equation}
    D\left(\mathbf{A}\right) = D_{l_2}\left(D_{l_1}\left(\text{flatten}\left(\mathbf{A}\right)\right)\right) \in \mathbb{R}^2,
    \label{eq:dhcp_detector}
\end{equation}
where ``flatten'' refers to reshaping the cross-modal attention into a one-dimensional vector, while $D_{l_1}$ and $D_{l_2}$ represent the first and second layers of the detector $D$, respectively. The first output corresponds to the non-hallucination score, and the second output corresponds to the hallucination score.

\begin{figure*}[t]
    \centering
    \includegraphics[width=0.95\linewidth]{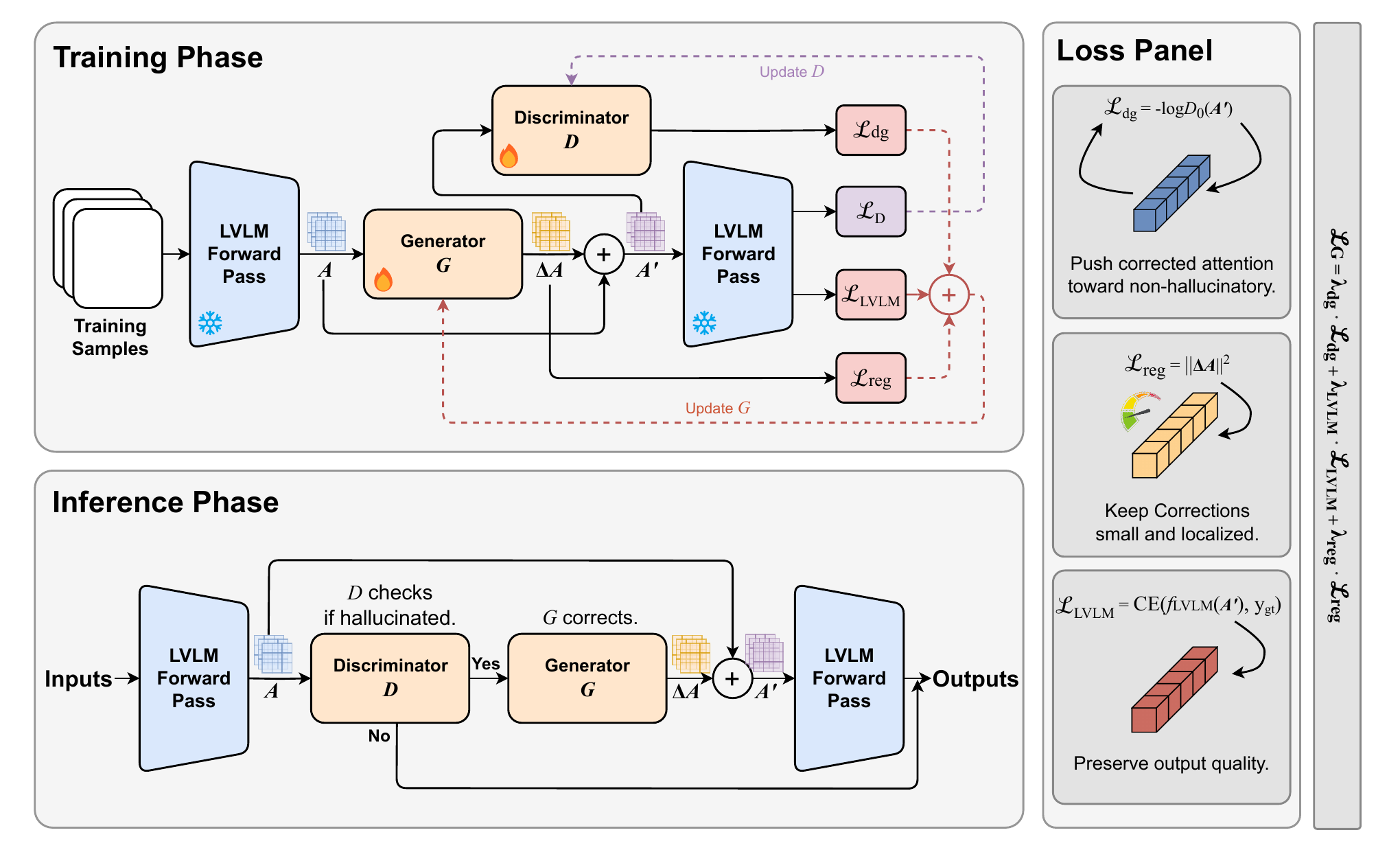}
    \vspace{-2em}
    \caption{The MHSA method pipeline. Discriminator $D$ is responsible for detecting hallucinations, while Generator $G$ is responsible for correcting cross-modal attention, forming a two-stage ``hallucination detection--mitigation'' framework.}
    \label{fig:method}
    % \vspace{-2em}
\end{figure*}

\subsection{Problem Formulation}
\label{sec:formulation}

While attention-based hallucination detection methods can effectively \emph{detect} hallucinations through cross-modal attention patterns, they do not provide the capability for hallucination \emph{mitigation}. We aim to develop a method that can directly \emph{correct} the attention patterns causing hallucinations.

Formally, given a cross-modal attention $\mathbf{A}$ that is detected as hallucinatory by the detector (\ie, $\arg\max D(\mathbf{A}) = 1$), our goal is to find an attention correction $\Delta\mathbf{A}$ such that the corrected attention:
\begin{equation}
    \mathbf{A}' = \mathbf{A} + \Delta\mathbf{A}
    \label{eq:corrected_attention}
\end{equation}
satisfies the following desiderata:
\begin{enumerate}
    \item[(1)] $\mathbf{A}'$ is no longer classified as hallucinatory by $D$, \ie, $\arg\max D(\mathbf{A}') = 0$
    \item[(2)] $\Delta\mathbf{A}$ is minimal to preserve the original semantics and avoid introducing new artifacts
    \item[(3)] The LVLM output generated with $\mathbf{A}'$ maintains or improves quality compared to the output with $\mathbf{A}$
\end{enumerate}

\subsection{MHSA: Attention Steering Framework}
\label{sec:mhsa_framework}

We propose MHSA, which employs a three-goal guided training strategy to learn the attention correction. The framework consists of a generator $G$ and a discriminator $D$ (the pre-trained hallucination detector).

\subsubsection{Generator $G$.}
\label{sec:generator}

The generator $G$ takes the original cross-modal attention $\mathbf{A}$ as input and produces the attention correction $\Delta\mathbf{A}$:
\begin{equation}
    \Delta\mathbf{A} = G(\mathbf{A}),
    \label{eq:generator}
\end{equation}
where $G$ is implemented as a three-layer MLP. The input attention tensor $\mathbf{A} \in \mathbb{R}^{L \times H \times N}$ is first flattened into a vector of dimension $d = L \cdot H \cdot N$. All hidden layers in the MLP have a dimension of 512, with the ReLU activation function applied between each layer, and the final output dimension is the same as the input dimension.

All linear layer weights are initialized with small values ($\mathcal{U}(-10^{-5}, 10^{-5})$) and zero bias, ensuring that the initial correction $\Delta\mathbf{A}$ is near-zero at the start of training. Since all input images are resized to a fixed resolution (see \cref{sec:preliminaries}), $N$ is constant across all samples, ensuring that $G$ has a fixed architecture that can be trained across the entire dataset. The corrected attention is then added to the original attention, as shown in \cref{eq:corrected_attention}.

We adopt a residual formulation $\mathbf{A}' = \mathbf{A} + \Delta \mathbf{A}$ rather than directly predicting $\mathbf{A}'$ from scratch, so that the generator only needs to learn a small offset from the original cross-modal attention. This residual design preserves the pretrained LVLM's attention structure as the starting point and confines the generator's role to localized refinement. Together with the regularization loss in \cref{eq:loss_reg}, it keeps $\mathbf{A}'$ close to $\mathbf{A}$, which we find empirically yields stable training and inference across all evaluated models and datasets.

\subsubsection{Discriminator $D$ (Hallucination Detector).}
\label{sec:discriminator}

The discriminator $D$ is the pre-trained hallucination detector~\cite{zhang2025dhcp}. During MHSA training, $D$ evaluates whether the original attention $\mathbf{A}$ exhibits hallucinatory patterns. The parameters of $D$ are \textbf{fine-tuned} throughout the training process with a tiny learning rate, allowing it to serve as a nearly fixed yet slightly flexible supervisory signal that guides the generator to produce corrections moving attention patterns away from hallucination-associated regions in the detector's learned feature space.

\subsection{Training Objectives}
\label{sec:objectives}

The MHSA framework is optimized with three complementary loss functions, corresponding to the three desiderata in \cref{sec:formulation}, as shown in \cref{fig:method}.

\textbf{Detector-Guided Loss ($\mathcal{L}_{\text{dg}}$).} Inspired by the adversarial training paradigm~\cite{goodfellow2014gan}, we design a detector-guided loss that drives the generator to produce corrections transforming hallucinated attention patterns into non-hallucinatory ones, as judged by the discriminator. For a hallucinatory sample with cross-modal attention $\mathbf{A}$:
\begin{equation}
    \mathcal{L}_{\text{dg}} = -\log D_0(\mathbf{A}') = -\log D_0(\mathbf{A} + G(\mathbf{A})),
    \label{eq:loss_adv}
\end{equation}
where $D_0(\cdot)$ denotes the probability assigned to the non-hallucinatory class (class 0) by the discriminator. Minimizing $\mathcal{L}_{\text{dg}}$ encourages the corrected attention to be classified as non-hallucinatory.

\textbf{Regularization Loss ($\mathcal{L}_{\text{reg}}$).} To prevent the generator from producing excessively large corrections that could disrupt the original attention semantics, we impose a regularization constraint on $\Delta\mathbf{A}$:
\begin{equation}
    \mathcal{L}_{\text{reg}} = \|\Delta\mathbf{A}\|_2^2 = \|G(\mathbf{A})\|_2^2.
    \label{eq:loss_reg}
\end{equation}
This loss ensures that the correction remains small and localized, preserving the overall structure of the original cross-modal attention while making targeted adjustments to hallucination-inducing patterns.

\textbf{LVLM Output Quality Loss ($\mathcal{L}_{\text{LVLM}}$).} To ensure that the corrected attention leads to high-quality outputs from the LVLM, we include a loss term that measures the impact of the attention correction on the LVLM's generation:
\begin{equation}
    \mathcal{L}_{\text{LVLM}} = \text{CE}(f_{\text{LVLM}}(\mathbf{A}'), y_{\text{gt}}),
    \label{eq:loss_llm}
\end{equation}
where $\text{CE}(\cdot, \cdot)$ denotes the cross-entropy loss, $f_{\text{LVLM}}(\mathbf{A}')$ represents the LVLM's output distribution when using the corrected attention $\mathbf{A}'$, and $y_{\text{gt}}$ is the ground-truth label. This loss ensures that the attention correction not only reduces hallucination signatures but also maintains or improves the actual output quality.

\textbf{Overall Objective.}
\label{sec:overall}

The total training objective for the generator $G$ is:
\begin{equation}
    \mathcal{L}_{\text{total}} = \lambda_{\text{dg}} \cdot \mathcal{L}_{\text{dg}} + \lambda_{\text{reg}} \cdot \mathcal{L}_{\text{reg}} + \lambda_{\text{LVLM}} \cdot \mathcal{L}_{\text{LVLM}},
    \label{eq:loss_total}
\end{equation}
where $\lambda_{\text{reg}}$ and $\lambda_{\text{LVLM}}$ are hyperparameters controlling the relative importance of the regularization and LVLM quality losses, respectively.

At the same time, we train the discriminator $D$ using the original cross-modal attention $\mathbf{A}$ of the current sample and its hallucination label as supervision:
\begin{equation}
    \mathcal{L}_{\text{d}} = -\left[y \log D_1(\mathbf{A}) + (1 - y) \log D_0(\mathbf{A})\right],
    \label{eq:loss_dhcp}
\end{equation}
where $y \in \{0, 1\}$ is the hallucination label ($y=1$ for hallucinatory, $y=0$ for non-hallucinatory), $D_1(\cdot)$ and $D_0(\cdot)$ denote the probabilities assigned to the hallucinatory and non-hallucinatory classes by the discriminator, respectively.

Only the generator $G$ and the discriminator $D$ (with a tiny learning rate) are trained; the LVLM backbone remains frozen throughout.

\subsection{Inference}
\label{sec:inference}

During inference, MHSA operates as illustrated in \cref{fig:method}:
\begin{enumerate}
    \item Run the LVLM forward pass and extract the cross-modal attention $\mathbf{A}$ from the first output token over visual tokens (\cref{eq:dhcp_dis}).
    \item Pass $\mathbf{A}$ through the pre-trained detector $D$ to determine whether the sample exhibits hallucinatory patterns.
    \item If hallucination is detected, apply the generator: $\Delta\mathbf{A} = G(\mathbf{A})$ and compute $\mathbf{A}' = \mathbf{A} + \Delta\mathbf{A}$.
    \item Replace the original cross-modal attention with $\mathbf{A}'$ in the LVLM to produce the corrected output.
\end{enumerate}

Since both $G$ and $D$ are lightweight MLPs (a two-layer MLP with hidden dimension 128 for $D$; a three-layer MLP with hidden dimension 512 for $G$), the additional computation introduced by MHSA is marginal compared to the LVLM's own inference cost involving billions of parameters. Importantly, MHSA does not modify any LVLM backbone parameters---only the external generator is trained offline.

\subsection{Extension to Token-Level Correction for Generative Tasks}
\label{sec:token_level}

For generative tasks such as image captioning, hallucinations may occur at specific tokens within the generated sequence rather than affecting the entire response. To handle this, we extend MHSA to operate at the token level.

Instead of using a single attention tensor derived from one output position (as in \cref{eq:dhcp_dis} for discriminative tasks), we consider the cross-modal attention for each individual generated token $m$:
\begin{equation}
    \mathbf{A}_m = \mathbf{A}^{(l,h)}_{q_m \to n}, \quad l = 1, \ldots, L, \; h = 1, \ldots, H, \; n = 1, \ldots, N,
    \label{eq:token_cma}
\end{equation}
where $q_m$ denotes the output position corresponding to the $m$-th generated token. Each $\mathbf{A}_m$ is a tensor of shape $(L, H, N)$, capturing the cross-modal attention pattern specific to the generation of that token.

To enable token-level hallucination detection, we upgrade the original sentence-level DHCP detector to a token-level detector. The attention training samples for the token-level detector are obtained as follows: we let the LVLM perform image captioning on training images and simultaneously extract the per-token cross-modal attention. For each generated caption, we detect all nouns in the generated text. For each noun, we check whether it appears in the CHAIR object whitelist~\cite{rohrbach2018chair}; if it does, we further verify whether it constitutes a hallucination (\ie, whether the object is absent from the ground-truth annotations). Nouns that are present in the whitelist and correctly describe objects in the image are labeled as \emph{non-hallucinated attention samples}, while those that are hallucinated are labeled as \emph{hallucinated attention samples}. The corresponding per-token cross-modal attention patterns are then used to train the token-level detector and the token-level generator $G$:
\begin{equation}
    \Delta\mathbf{A}_m = G(\mathbf{A}_m).
    \label{eq:token_generator}
\end{equation}

During generative inference, MHSA operates at the decoding steps covered by the CHAIR protocol, before the corresponding token is sampled. At such a step $m$, we first run a standard LVLM forward pass to obtain the cross-modal attention $\mathbf{A}_m$ for the candidate token at position $m$. The token-level discriminator $D$ then evaluates whether $\mathbf{A}_m$ exhibits a hallucinatory pattern. If so, the generator produces a correction $\Delta \mathbf{A}_m = G(\mathbf{A}_m)$, and the corrected attention $\mathbf{A}_m' = \mathbf{A}_m + \Delta \mathbf{A}_m$ replaces $\mathbf{A}_m$ in the LVLM forward pass to recompute the output logits, from which the token at position $m$ is then sampled.

\section{Experiment}
\label{sec:experiment}

In this section, we evaluate the effectiveness of MHSA on both discriminative and generative hallucination mitigation tasks. We first describe the experimental setup (\cref{sec:setup}), then present the main results on discriminative POPE tasks (\cref{sec:main_results}), followed by dataset generalization (\cref{sec:dataset_gen}), model generalization (\cref{sec:model_gen}), cross-dataset out-of-distribution generalization (\cref{sec:ood_gen}), generative captioning results (\cref{sec:gen_results}), and ablation studies (\cref{sec:ablation}).

\subsection{Experimental Setup}
\label{sec:setup}

\subsubsection{Models.}
We conduct experiments on three representative LVLMs to demonstrate the generality of MHSA: Qwen2.5-VL-7B~\cite{bai2025qwen25vl}, InternVL2-8B~\cite{chen2024internvl}, and LLaVA-v1.5-7B~\cite{liu2024visual, liu2023improved}. For models with variable visual token numbers, Qwen2.5-VL and InternVL have visual token sizes set to 144 and 256, respectively. For each model, the hallucination detector $D$ is pre-trained on the corresponding training split's cross-modal attention data following the protocol described in~\cite{zhang2025dhcp}. The generator $G$ is a three-layer MLP with hidden dimension 512.

\subsubsection{Datasets.}
For POPE tasks, we evaluate on four datasets: \textbf{MSCOCO}~\cite{lin2014coco}, \textbf{Objects365}~\cite{objects365}, \textbf{OpenImagesV7}~\cite{DBLP:journals/corr/abs-1811-00982}, and \textbf{ImageNet}~\cite{deng2009imagenet}. POPE questions are generated following the POPE framework~\cite{li2023evaluating} with three evaluation clusters: Random, Popular, and Adversarial. 
For captioning tasks, we evaluate on two datasets: \textbf{MSCOCO}~\cite{lin2014coco} and \textbf{Flickr30k}~\cite{plummer2015flickr30k}, assessed using the CHAIR metric~\cite{rohrbach2018chair}.

\subsubsection{Evaluation Metrics.}
For discriminative tasks, we report Accuracy, Precision, Recall, F1-score, and Yes Ratio. The F1-score is computed for the ``Yes'' label following the POPE convention. In some cases, overtrained models may produce invalid outputs which cannot be assigned to any of the standard classification categories, \ie, TP, TN, FP, or FN. To address this, we report accuracy; these unresolved samples are included in the denominator. For generative tasks, we report CHAIR$_s$ (lower is better), CHAIR$_i$ (lower is better), and Recall (higher is better).

\subsubsection{Implementation Details.}
The MHSA generator $G$ is implemented as a three-layer MLP with hidden dimension 512. The flattened attention dimension $d = L \cdot H \cdot N$ differs across models: $d = 28 \times 28 \times 144 = 112{,}896$ for Qwen2.5-VL-7B, $d = 32 \times 32 \times 256 = 262{,}144$ for InternVL2-8B, and $d = 32 \times 32 \times 576 = 589{,}824$ for LLaVA-v1.5-7B. The linear layer weight of $G$ is initialized with small weights ($\mathcal{U}(-10^{-5}, 10^{-5})$), and the bias is initialized to zero.

The hallucination detector $D$ is pre-trained following the protocol in~\cite{zhang2025dhcp}. The loss weights are set to $\lambda_{\text{LVLM}} = 1$ and $\lambda_{\text{dg}} = 0.01$ by default. $\lambda_{\text{reg}}$ differs between models. The attention correction is applied across all LLM layers by default.

For the captioning task, we adopt an offline training mode: per-token cross-modal attention tensors are pre-extracted and stored, and the generator $G$ is trained without loading the LVLM, using only the adversarial loss and regularization loss ($\mathcal{L}_{\text{LVLM}}$ is not applicable in offline mode).

\subsection{Main Results on POPE-COCO}
\label{sec:main_results}

We begin by presenting detailed POPE results with Qwen2.5-VL-7B on MSCOCO, as shown in \cref{tab:main_coco}. This serves as the primary evidence for MHSA's effectiveness.

\begin{table}[t]
\centering
\caption{POPE results on MSCOCO (Qwen2.5-VL-7B, $N$=3000). Best results in \textbf{bold}.}
\label{tab:main_coco}
\resizebox{\columnwidth}{!}{
\begin{tabular}{c|ccccc}
    \toprule
    Method & Accuracy & Precision & Recall & F1 & Yes\% \\
    \midrule
    Baseline & 86.83 & \textbf{95.27} & 77.62 & 85.55 & 40.9 \\
    MHSA & \textbf{92.77} & 92.54 & \textbf{93.40} & \textbf{92.97} & 50.5 \\
    \midrule
    $\Delta$ & +5.94 & $-$2.73 & +15.78 & +7.42 & +9.6 \\
    \bottomrule
\end{tabular}
}
\end{table}

MHSA yields a substantial improvement of +7.42 F1 points. Notably, the Yes Ratio shifts from 40.9\% to 50.5\%, close to the balanced 50\%, indicating that the attention correction alleviates the model's tendency to over-predict ``No'' answers. The Recall improvement of +15.78 points is particularly significant, demonstrating that the corrected attention enables the model to correctly recognize present objects that were previously missed. While Precision decreases slightly ($-$2.73), the overall F1 improvement confirms a net gain in discriminative quality.

\subsection{Dataset Generalization}
\label{sec:dataset_gen}

To evaluate whether MHSA generalizes across different visual domains, we fix the model as Qwen2.5-VL-7B and test it on additional POPE benchmarks. For each dataset, the detector and generator are trained and tested on the same dataset. \cref{tab:dataset_gen} presents the results.

\begin{table}[t]
\centering
\caption{Dataset generalization on POPE (Qwen2.5-VL-7B). All experiments use the same hyperparameter configuration as the COCO experiments without dataset-specific tuning. Best results in \textbf{bold}.}
\label{tab:dataset_gen}
\begin{tabular}{c|c|ccccc}
    \toprule
    Dataset & Method & Acc & Prec & Recall & F1 & Yes\% \\
    \midrule
    \multirow{2}{*}{Objects365} & Baseline & 83.93 & \textbf{92.84} & 73.52 & 82.06 & 39.6 \\
    & MHSA & \textbf{91.23} & 91.87 & \textbf{90.46} & \textbf{91.16} & 49.2 \\
    \midrule
    \multirow{2}{*}{ImageNet} & Baseline & 82.80 & \textbf{96.33} & 68.20 & 79.86 & 35.4 \\
    & MHSA & \textbf{86.67} & 84.25 & \textbf{90.20} & \textbf{87.12} & 53.5 \\
    \bottomrule
\end{tabular}
\end{table}

MHSA achieves consistent improvements on Objects365 ($\Delta$F1=+9.10) and ImageNet ($\Delta$F1=+7.26), both without any dataset-specific hyperparameter tuning. Detailed per-category results are provided in the Appendix.

\subsection{Model Generalization}
\label{sec:model_gen}

To demonstrate that MHSA generalizes across different model architectures, we evaluate InternVL2-8B and LLaVA-v1.5-7B across MSCOCO, Objects365, and OpenImagesV7. \cref{tab:model_gen} presents the results.

\begin{table}[t]
\centering
\caption{Model generalization on POPE across three datasets. Best results in \textbf{bold}.}
\label{tab:model_gen}

\centering
\subcaption{InternVL2-8B}
\label{tab:model_gen_internvl}
\begin{tabular}{c|c|ccccc}
    \toprule
    Dataset & Method & Acc & Prec & Recall & F1 & Yes\% \\
    \midrule
    \multirow{2}{*}{COCO} & Baseline & 87.07 & \textbf{90.86} & 82.54 & 86.50 & 45.6 \\
    & MHSA & \textbf{93.87} & 90.16 & \textbf{98.54} & \textbf{94.16} & 54.9 \\
    \midrule
    \multirow{2}{*}{Obj365} & Baseline & 83.97 & \textbf{89.60} & 76.90 & 82.77 & 43.0 \\
    & MHSA & \textbf{90.53} & 89.19 & \textbf{92.28} & \textbf{90.71} & 51.8 \\
    \midrule
    \multirow{2}{*}{OpenImg} & Baseline & 79.53 & 74.77 & 89.77 & 81.58 & 60.6 \\
    & MHSA & \textbf{83.73} & \textbf{75.74} & \textbf{99.74} & \textbf{86.10} & 66.5 \\
    \bottomrule
\end{tabular}

\hfill

\centering
\subcaption{LLaVA-v1.5-7B}
\label{tab:model_gen_llava}
\begin{tabular}{c|c|ccccc}
    \toprule
    Dataset & Method & Acc & Prec & Recall & F1 & Yes\% \\
    \midrule
    \multirow{2}{*}{COCO} & Baseline & 85.57 & \textbf{91.49} & 78.55 & 84.53 & 43.1 \\
    & MHSA & \textbf{92.10} & 90.70 & \textbf{93.89} & \textbf{92.27} & 52.0 \\
    \midrule
    \multirow{2}{*}{Obj365} & Baseline & 82.57 & \textbf{90.56} & 72.77 & 80.69 & 40.2 \\
    & MHSA & \textbf{90.37} & 85.66 & \textbf{98.12} & \textbf{91.47} & 57.0 \\
    \midrule
    \multirow{2}{*}{OpenImg} & Baseline & 78.50 & 72.69 & 90.09 & 80.46 & 60.9 \\
    & MHSA & \textbf{81.07} & \textbf{74.20} & \textbf{96.05} & \textbf{83.72} & 63.3 \\
    \bottomrule
\end{tabular}

\end{table}

MHSA improves F1 across all six model-dataset combinations, with gains ranging from +3.26 (LLaVA on OpenImages) to +10.78 (LLaVA on Objects365). InternVL2-8B on COCO achieves the highest F1 of 94.16 ($\Delta$F1=+7.66). These results confirm that the learned attention correction generalizes well across diverse LVLM architectures.

\subsection{Cross-Dataset Out-of-Distribution Generalization}
\label{sec:ood_gen}

A critical question is whether the attention correction patterns learned by MHSA can transfer across data distributions. We train MHSA on one dataset and test it on another, constructing a cross-dataset generalization matrix. \cref{tab:ood} presents the results in terms of F1 scores, with in-domain results on the diagonal and out-of-distribution (OOD) results off the diagonal.

\begin{table}[t]
\centering
\caption{Cross-dataset OOD generalization (F1 scores). Rows = training dataset, columns = test dataset. Diagonal entries (bold) = in-domain.}
\label{tab:ood}
\subcaption{InternVL2-8B}
\begin{tabular}{l|cc}
    \toprule
    Datasets & COCO & Obj365 \\
    \midrule
    COCO   & \textbf{94.16} & 90.57 \\
    Obj365 & 95.77 & \textbf{90.71} \\
    \textit{Baseline} & \textit{86.50} & \textit{82.77} \\
    \bottomrule
\end{tabular}

\subcaption{LLaVA-v1.5-7B}
\begin{tabular}{l|cc}
    \toprule
    Datasets & COCO & Obj365 \\
    \midrule
    COCO   & \textbf{92.27} & 89.50 \\
    Obj365 & 93.97 & \textbf{91.47} \\
    \textit{Baseline} & \textit{84.53} & \textit{80.69} \\
    \bottomrule
\end{tabular}

\subcaption{Qwen2.5-VL-7B}
\begin{tabular}{l|cc}
    \toprule
    Datasets & COCO & Obj365 \\
    \midrule
    COCO   & \textbf{92.97} & 91.02 \\
    Obj365 & 93.48 & \textbf{91.16} \\
    \textit{Baseline} & \textit{85.55} & \textit{82.06} \\
    \bottomrule
\end{tabular}

\end{table}

All OOD entries surpass the corresponding baselines, demonstrating that the attention correction learned by MHSA transfers well across data distributions. Remarkably, InternVL2-8B trained on Objects365 achieves 95.77 F1 on COCO, exceeding its in-domain performance of 94.16, suggesting that the learned correction captures fundamental hallucination patterns that generalize beyond the training data. Similarly, for Qwen2.5-VL-7B, the OOD setting (Obj365$\to$COCO: 93.48) outperforms the in-domain setting (COCO$\to$COCO: 92.97). These findings strongly support that MHSA learns generalizable attention correction rather than dataset-specific artifacts.

\subsection{Generative Hallucination Mitigation (Captioning)}
\label{sec:gen_results}
\begin{table}[t]
\centering
\caption{Caption Generation Results (CHAIR, Qwen2.5-VL-7B)}
\label{tab:caption_generation_results}
\begin{tabular}{llccc}
\toprule
Dataset & Method & CHAIRi $\downarrow$ & CHAIRs $\downarrow$ & Recall $\uparrow$ \\
\midrule
\multirow{3}{*}{Flickr30k}
& Baseline & 16.43 & 37.50 & 86.63 \\
& MHSA & 9.20 & 21.00 & 83.28 \\
& $\Delta$ & -44.0\% & -44.0\% & -3.9\% \\
\midrule
\multirow{3}{*}{COCO}
& Baseline & 5.68 & 21.00 & 56.42 \\
& MHSA & 5.19 & 18.00 & 55.14 \\
& $\Delta$ & -8.6\% & -14.3\% & -2.3\% \\
\bottomrule
\end{tabular}
\end{table}
For generative tasks, we apply the token-level MHSA extension (\cref{sec:token_level}) to image captioning. The generator $G_{\text{token}}$ is trained offline on pre-extracted per-token attention tensors. \cref{tab:caption_generation_results} presents the results with Qwen2.5-VL-7B on Flickr30k and MSCOCO.

On Flickr30k, MHSA reduces both CHAIR$_i$ and CHAIR$_s$ by 44\% with only a 3.9\% drop in Recall, demonstrating a highly favorable trade-off between hallucination suppression and content preservation. On MSCOCO, where the baseline hallucination rate is already low, MHSA still achieves meaningful reductions (CHAIR$_i$: $-$8.6\%, CHAIR$_s$: $-$14.3\%) with minimal Recall loss ($-$2.3\%). These results demonstrate that the token-level attention correction effectively extends MHSA from discriminative to generative tasks.

\begin{figure}[t]
\centering
\includegraphics[width=\linewidth]{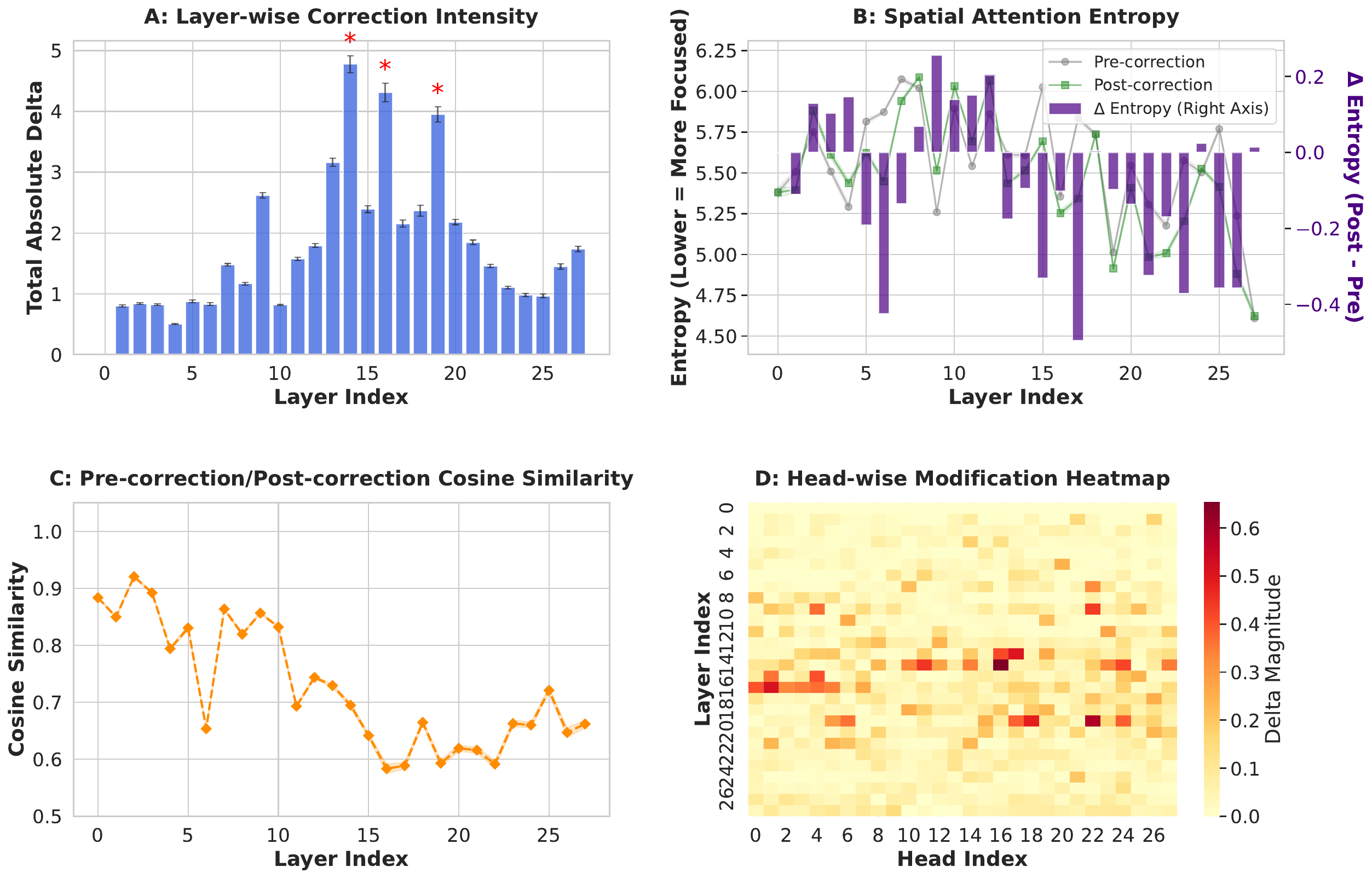}
\caption{Statistical analysis of attention modifications between pre-correction (hallucinated) and post-correction (factual) states. (A) Layer-wise correction intensity, measured by the total absolute delta of attention weights. Red asterisks denote the top three most modified layers. (B) Spatial attention entropy before (gray) and after (green) correction. Purple bars (right axis) represent the change in entropy ($\Delta$ Entropy). Lower entropy values indicate a higher degree of spatial focus. (C) Layer-wise cosine similarity between pre- and post-correction attention maps. (D) Heatmap detailing the absolute modification magnitude (delta) for each specific attention head. Error bars and shaded regions in panels A, B, and C represent the standard error of the mean (SEM).}
\label{fig:chart}
\end{figure}

\subsection{Ablation Studies}
\label{sec:ablation}
\begin{table}[htbp]
\centering
\caption{Ablation study on loss function combinations.}
\label{tab:ablation}
\begin{tabular}{lcccc}
\hline
Setting & $\mathcal{L}_{\text{LVLM}}$ & $\mathcal{L}_{\text{dg}}$ & $\mathcal{L}_{\text{reg}}$ & F1 \\
\hline
Baseline &  &  &  & 85.55 \\
w/o $\mathcal{L}_{\text{dg}}$ & w/ & w/o & w/ & 88.51 \\
w/o $\mathcal{L}_{\text{LVLM}}$ & w/o & w/ & w/ & 84.18 \\
w/o $\mathcal{L}_{\text{reg}}$ & w/ & w/ & w/o & 89.60 \\
$\mathcal{L}_{\text{reg}}$ only & w/o & w/o & w/ & 83.11 \\
MHSA (full) & w/ & w/ & w/ & \textbf{92.97} \\
\hline
\end{tabular}
\end{table}
We conduct ablation studies on Qwen2.5-VL-7B using POPE-COCO to analyze the contribution of each component in MHSA. \cref{tab:ablation} presents the results for different combinations of loss functions.

Removing any single loss term leads to a clear performance drop, and using $\mathcal{L}_{\text{reg}}$ alone provides little benefit. The full combination achieves the best F1, confirming that all three losses are effective and complementary.

\begin{figure}[t]
\centering
\includegraphics[width=\linewidth]{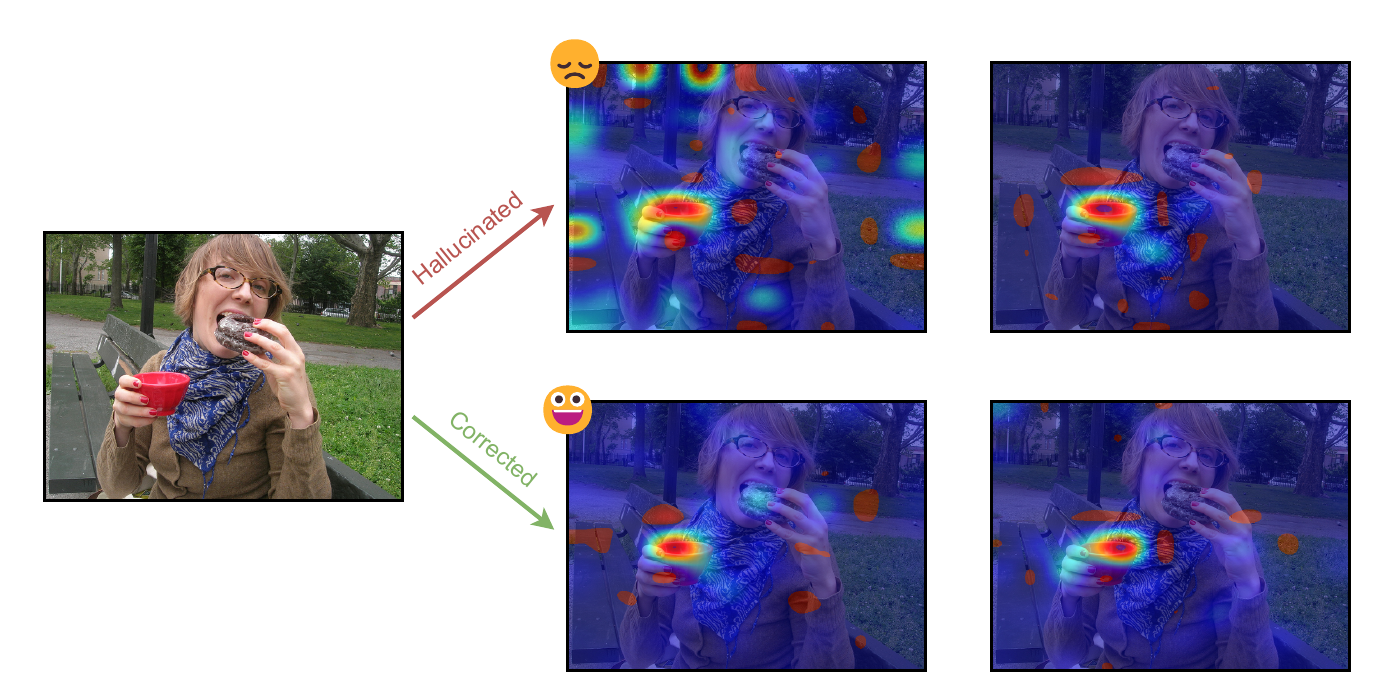}
\caption{Attention visualization before and after MHSA correction of the POPE discriminative task.}
\label{fig:pope_attn}
\end{figure}

\begin{figure}[t]
\centering
\includegraphics[width=\columnwidth]{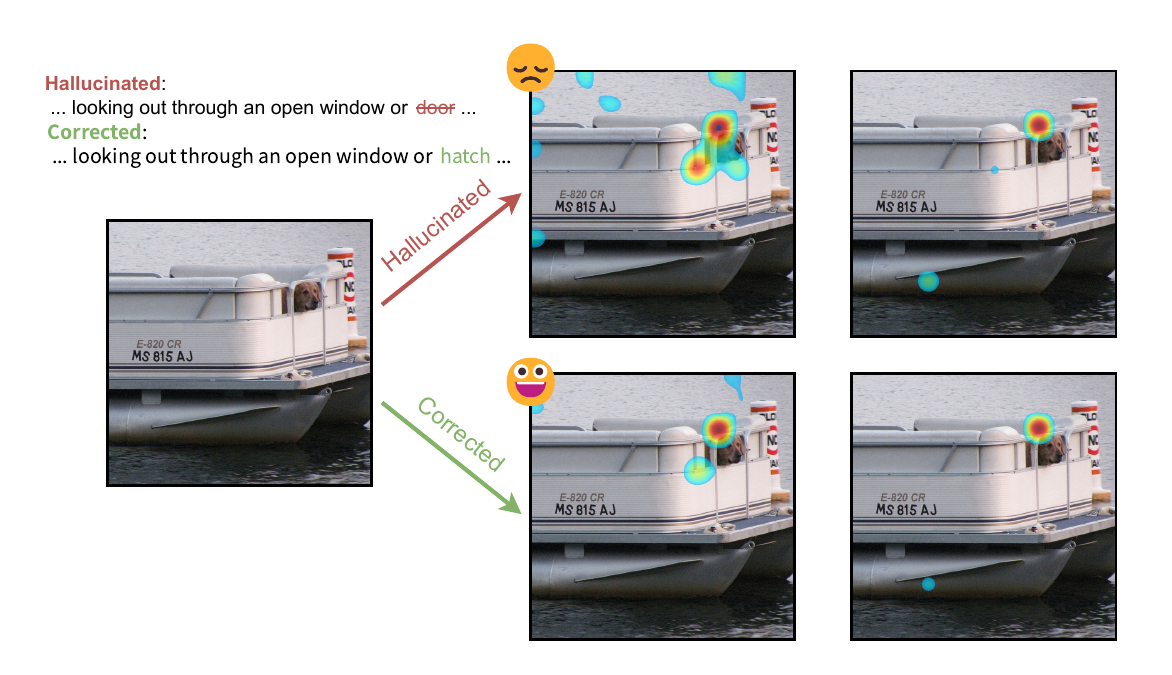}
\caption{Attention visualization before and after MHSA correction of the COCO generative task.}
\label{fig:caption_attn}
\end{figure}

\subsection{Mechanistic Analysis of Attention Correction}

\Cref{fig:chart} provides a statistical characterization of the modifications induced by MHSA in cross-modal attention. The correction is not distributed uniformly across the network; instead, it is concentrated in a limited set of intermediate layers, where the total absolute change in attention is the largest and the cosine similarity between pre-correction and post-correction attention shows the greatest reduction. At the same time, post-correction attention entropy is generally lower than pre-correction attention entropy, indicating that MHSA makes the spatial distribution of attention more concentrated. The head-wise heatmap further shows that these modifications are sparse, with substantial updates confined to only a small subset of layer-head pairs. Taken together, these observations suggest that MHSA functions as a targeted correction mechanism that intervenes primarily at the layers and heads where hallucination-related cross-modal misalignment is most pronounced, rather than broadly perturbing the full attention structure.

\Cref{fig:pope_attn,fig:caption_attn} provide qualitative examples that are consistent with the statistical evidence in \cref{fig:chart}. In the POPE discriminative example, the hallucinated prediction is associated with attention that is diffuse and partially misaligned, whereas after correction, the attention shifts toward the queried object itself, enabling a more visually grounded judgment. A similar pattern appears in the captioning example: before correction, the generated token is supported by attention assigned to an irrelevant region, whereas after correction, the attention is redirected to the visually relevant region, resulting in a more faithful description. These examples complement the statistical results and further support the conclusion that MHSA mitigates hallucination by selectively improving the spatial grounding of cross-modal attention.
\section{Conclusion}
\label{sec:conclusion}

In this paper, we propose MHSA (Mitigating Hallucinations via Steered Attention), a lightweight framework that learns sample-adaptive corrections of cross-modal attention patterns to mitigate hallucinations in large vision-language models. Unlike existing heuristic-based attention manipulation methods, MHSA trains a lightweight MLP generator guided by a pre-trained hallucination detector through adversarial supervision, without modifying any LVLM backbone parameters. Extensive experiments on three representative LVLMs (Qwen2.5-VL-7B, InternVL2-8B, and LLaVA-v1.5-7B) across multiple POPE benchmarks demonstrate consistent improvements in F1-score, with strong dataset, model, and cross-dataset out-of-distribution generalization. On generative captioning tasks, MHSA substantially reduces CHAIR scores. Future work will explore more expressive generator architectures, broader model coverage, and synergistic combinations with complementary mitigation strategies.

% \begin{acks}
% \end{acks}

\clearpage
\newpage
\bibliographystyle{ACM-Reference-Format}
\bibliography{sample-base,mhsa-refs}

\clearpage
\newpage
\appendix
\section{Inference Efficiency}
\label{sec:appendix_efficiency}

We benchmark the MHSA pipeline on POPE-COCO (1,000 samples, Qwen2.5-VL-7B, NVIDIA H20). Since MHSA only corrects the 12.3\% of samples detected as hallucinated, the vast majority (87.7\%) incur no correction overhead, resulting in an amortized mean latency increase of only 0.43$\times$.

\begin{table}[!htbp]
\centering
\caption{Overall inference efficiency (POPE-COCO, Qwen2.5-VL-7B).}
\label{tab:appendix_efficiency}
\begin{tabular}{lccc}
    \toprule
     & Baseline & MHSA & $\Delta$ \\
    \midrule
    Avg latency (ms) & 113.1 & 161.2 & $+0.43\times$ \\
    Throughput (samples/s) & 8.84 & 6.20 & $-0.30\times$ \\
    \bottomrule
\end{tabular}
\end{table}

\begin{table}[!htbp]
\centering
\caption{MHSA latency breakdown. 87.7\% of samples skip correction and run close to baseline speed.}
\label{tab:appendix_breakdown}
\begin{tabular}{lccc}
    \toprule
    Sample Type & Ratio & Avg (ms) & Median (ms) \\
    \midrule
    Non-Halluc.\ (no correction) & 87.7\% & 115.1 & 114.9 \\
    Hallucinated (corrected)     & 12.3\% & 486.4 & 205.5 \\
    \midrule
    All & 100\% & 161.2 & 115.1 \\
    \bottomrule
\end{tabular}
\end{table}

This $+0.4\times$ overhead benefits from the two-stage detect-then-correct design: the lightweight detector first screens all samples, and the more expensive MHSA generator is invoked only for the small fraction flagged as hallucinated. By contrast, contrastive decoding methods such as VCD~\cite{leng2023vcd} and ICD~\cite{wang2024icd} require at least two full forward passes for \emph{every} sample, incurring a fixed ${\sim}{+1\times}$ additional latency cost regardless of whether a given sample is hallucinated. Attention-based methods such as OPERA~\cite{huang2023opera} further introduce backtracking overhead during generation. Furthermore, the current $+0.4\times$ figure is measured on POPE ~\cite{li2023evaluating, lovenia2023nope} that employs adversarial negative sampling to deliberately maximize hallucination-prone scenarios, resulting in an elevated hallucination rate (\eg, 12.3\% on COCO as detected by DHCP). In real-world deployments, where queries are not adversarially constructed, the proportion of samples triggering MHSA correction would be substantially lower, driving the amortized overhead closer to $+0\times$---while contrastive decoding methods would still incur a constant ${+1\times}$ additional cost on every sample.

\section{Model Architecture Details}
\label{sec:appendix_arch}
\cref{tab:appendix_arch} lists the attention configurations for each LVLM. \cref{tab:appendix_gd} details the lightweight generator $G$ and detector $D$.
\begin{table}[!htbp]
\centering
\caption{Attention configurations for each LVLM. $L$: number of transformer layers; $H$: number of attention heads per layer; $N$: number of visual tokens; $d = L \cdot H \cdot N$: total number of cross-modal attention entries used as input to the generator and detector.}
\label{tab:appendix_arch}
\begin{tabular}{lcccc}
    \toprule
    Model & $L$ & $H$ & $N$ & $d = L \cdot H \cdot N$ \\
    \midrule
    Qwen2.5-VL-7B & 28 & 28 & 144 & 112,896 \\
    InternVL2-8B  & 32 & 32 & 256 & 262,144 \\
    LLaVA-v1.5-7B & 32 & 32 & 576 & 589,824 \\
    \bottomrule
\end{tabular}
\end{table}

\begin{table}[!htbp]
\centering
\caption{Generator $G$ and Detector $D$ architecture. $d$ is model-specific (\cref{tab:appendix_arch}).}
\label{tab:appendix_gd}
\begin{tabular}{lcccc}
    \toprule
     & Input & Hidden & Output & Structure \\
    \midrule
    $G$ & $d$ & 512 & $d$ & Lin-ReLU-Lin-ReLU-Lin \\
    $D$ & $d$ & 128 & 2 & LN-Lin-ReLU-Lin \\
    \bottomrule
\end{tabular}
\end{table}

\section{Training Hyperparameters}
\label{sec:appendix_hyper}

\cref{tab:appendix_hyperparams} lists all training hyperparameters. All configurations use weight decay = 1e-4 and train for 1 epoch.

\begin{table}[!htbp]
\centering
\caption{Training hyperparameters for MHSA. lr$_G$: generator learning rate; lr$_D$: discriminator learning rate; $\lambda_{\text{LVLM}}$: LVLM quality loss weight; $\lambda_{\text{dg}}$: detector-guided loss weight; $\lambda_{\text{reg}}$: regularization loss weight.}
\label{tab:appendix_hyperparams}
\resizebox{\columnwidth}{!}{
\begin{tabular}{c|c|ccccccc}
    \toprule
    Task & Model & lr$_G$ & lr$_D$ & $\lambda_{\text{LVLM}}$ & $\lambda_{\text{dg}}$ & $\lambda_{\text{reg}}$ & Epochs & Batch \\
    \midrule
    POPE & Qwen2.5-VL & 1e-4 & 1e-5 & 1.0 & 0.01 & 1e-4 & 1 & 16 \\
    POPE & LLaVA-v1.5 & 1e-4 & 1e-5 & 1.0 & 0.01 & 5e-4 & 1 & 8 \\
    POPE & InternVL2 & 1e-3 & 1e-4 & 1.0 & 0.01 & 1e-4 & 1 & 8 \\
    Caption & Qwen2.5-VL & 1e-3 & 1e-7 & 0.0 & 0.5 & 0.01 & 1 & 32 \\
    \bottomrule
\end{tabular}
}
\end{table}

\section{Dataset and Training Statistics}
\label{sec:appendix_data}

Each POPE dataset contains 3,000 Yes/No questions (1,000 per category: Popular, Random, Adversarial; 500 positive + 500 negative each). For each model$\times$dataset, we extract per-layer cross-modal attention and split 80/20 by question ID (seed=42). DHCP classifies entries into Class~0/1 (non-hallucination) and Class~2/3 (hallucination). MHSA training oversamples: all Class~2/3 retained, Class~0/1 subsampled to half of Class~2+3 total.

\begin{table}[!htbp]
\centering
\caption{MHSA training set after oversampling. Cls\,0/1 = non-halluc., Cls\,2/3 = halluc.}
\label{tab:appendix_gan}
\resizebox{\columnwidth}{!}{
\begin{tabular}{c|c|ccccc}
    \toprule
    Model & Data & Cls\,0 & Cls\,1 & Cls\,2 & Cls\,3 & Total \\
    \midrule
    \multirow{3}{*}{Qwen2.5-VL} & COCO & 20.8k & 20.8k & 36.1k & 5.5k & 83.3k \\
     & Obj365 & 29.3k & 29.3k & 47.4k & 11.3k & 117.4k \\
     & OImages & 35.5k & 35.5k & 22.1k & 48.9k & 142.0k \\
    \midrule
    \multirow{3}{*}{LLaVA-v1.5} & COCO & 22.3k & 22.3k & 34.6k & 10.0k & 89.0k \\
     & Obj365 & 31.3k & 31.3k & 47.3k & 15.4k & 125.2k \\
     & OImages & 40.5k & 40.5k & 20.9k & 60.1k & 161.9k \\
    \midrule
    \multirow{3}{*}{InternVL2} & COCO & 21.4k & 21.4k & 29.7k & 13.2k & 85.7k \\
     & Obj365 & 30.3k & 30.3k & 42.6k & 18.0k & 121.2k \\
     & OImages & 38.8k & 38.8k & 18.2k & 59.3k & 155.1k \\
    \bottomrule
\end{tabular}
}
\end{table}

\section{Hyperparameter Sensitivity Analysis}
\label{sec:appendix_sensitivity}

We conduct a comprehensive hyperparameter sensitivity study on POPE-COCO with Qwen2.5-VL-7B. For each hyperparameter, we vary it while keeping all others at their default values, and report Accuracy, Precision, Recall, and F1 across multiple random seeds. Box plots show the distribution; diamonds denote means; asterisks indicate statistical significance ($^*p{<}0.05$, $^{**}p{<}0.01$, $^{***}p{<}0.001$).

\begin{figure}[h]
\centering
\includegraphics[width=\columnwidth]{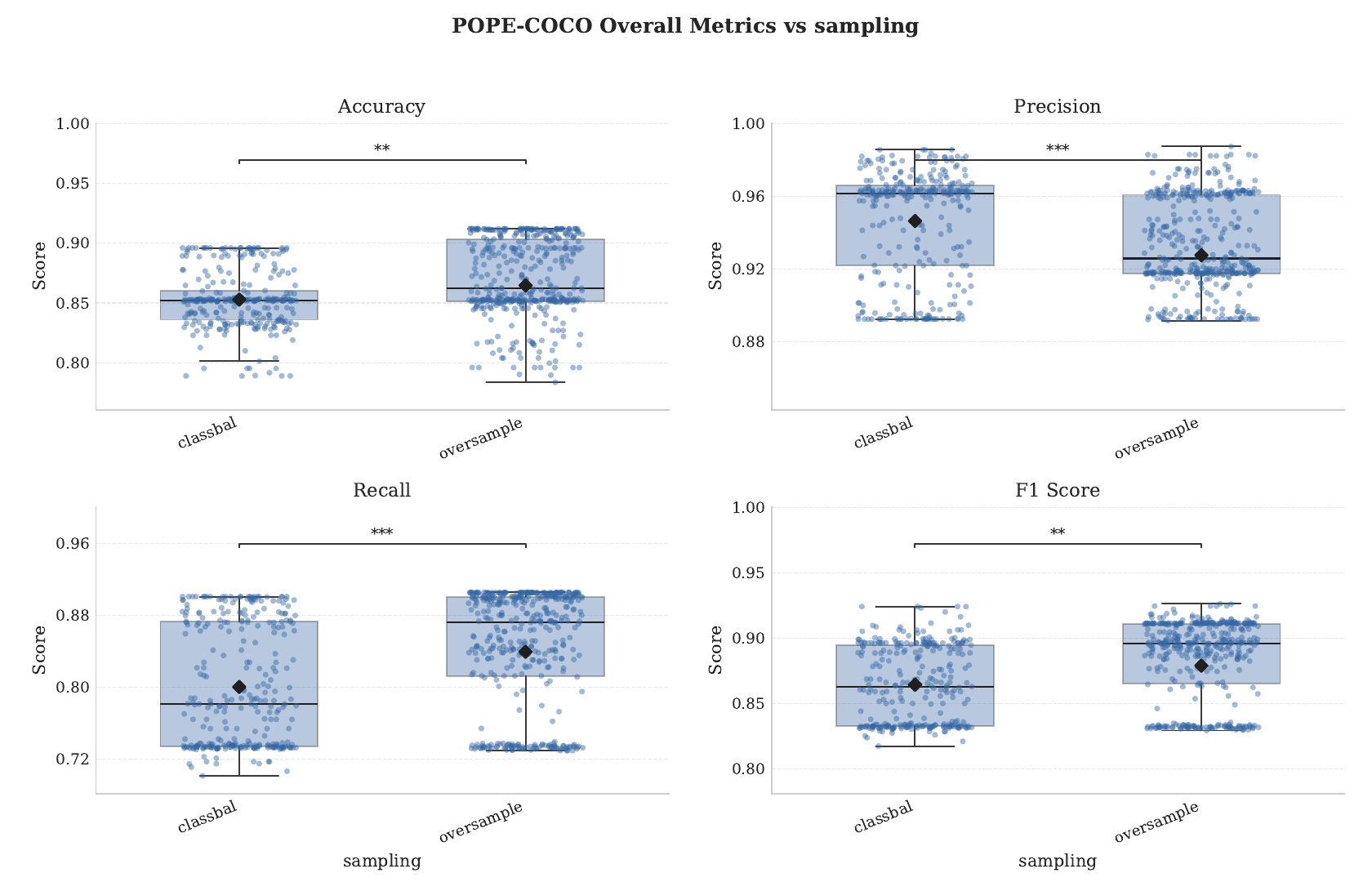}
\caption{POPE-COCO metrics vs.\ sampling strategy.}
\label{fig:supp_sampling}
\end{figure}

\paragraph{Sampling strategy.} \cref{fig:supp_sampling} compares class-balanced vs.\ oversampling. Oversampling yields significantly higher Recall and F1, confirming its effectiveness.

\begin{figure}[h]
\centering
\includegraphics[width=\columnwidth]{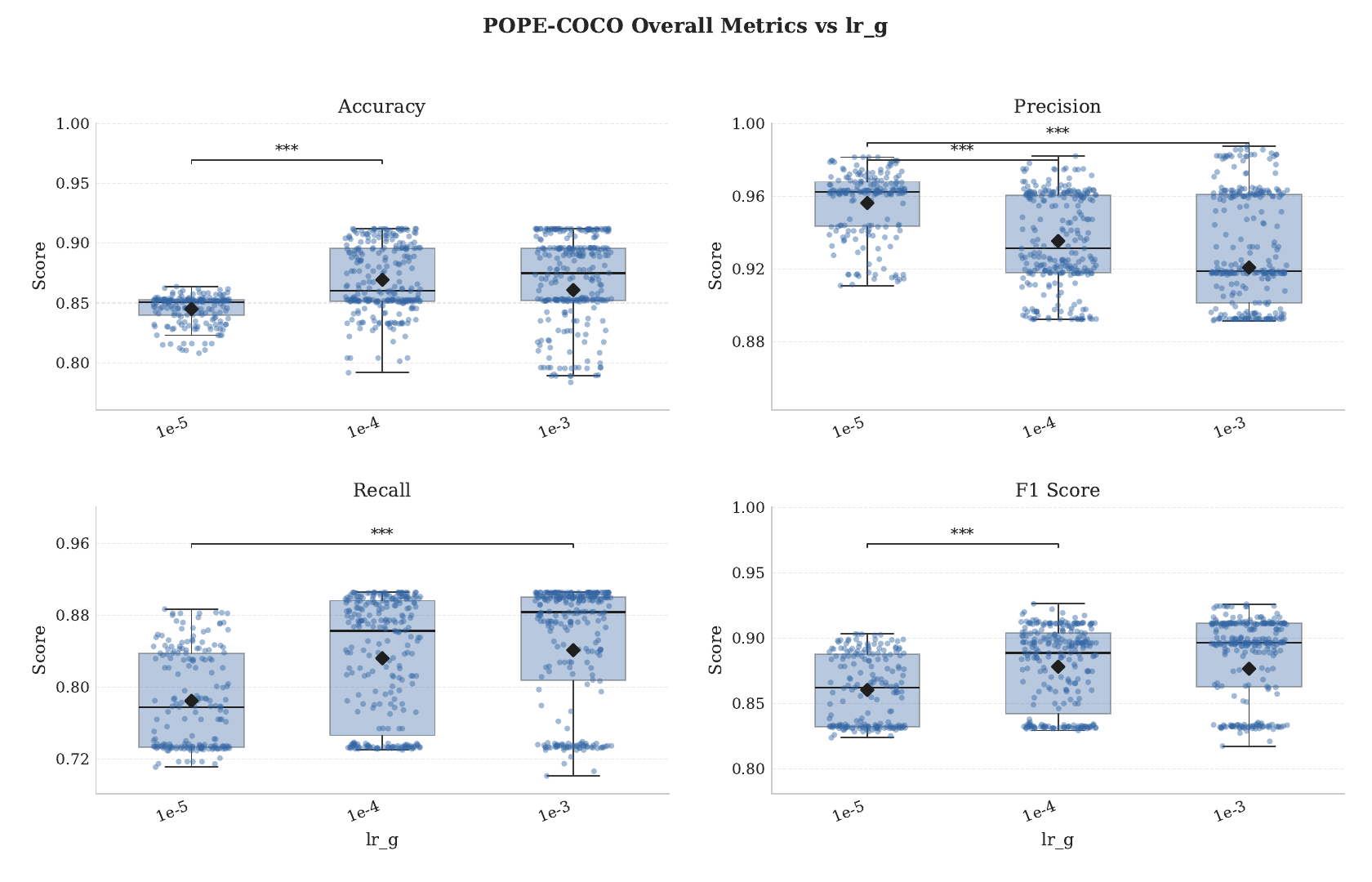}
\caption{POPE-COCO metrics vs.\ lr$_G$.}
\label{fig:supp_lr_g}
\end{figure}

\paragraph{Generator learning rate (lr$_G$).} \cref{fig:supp_lr_g} shows that lr$_G$=1e-4 achieves the best F1 and Accuracy. Too small (1e-5) under-corrects; too large (1e-3) increases variance.

\begin{figure}[h]
\centering
\includegraphics[width=\columnwidth]{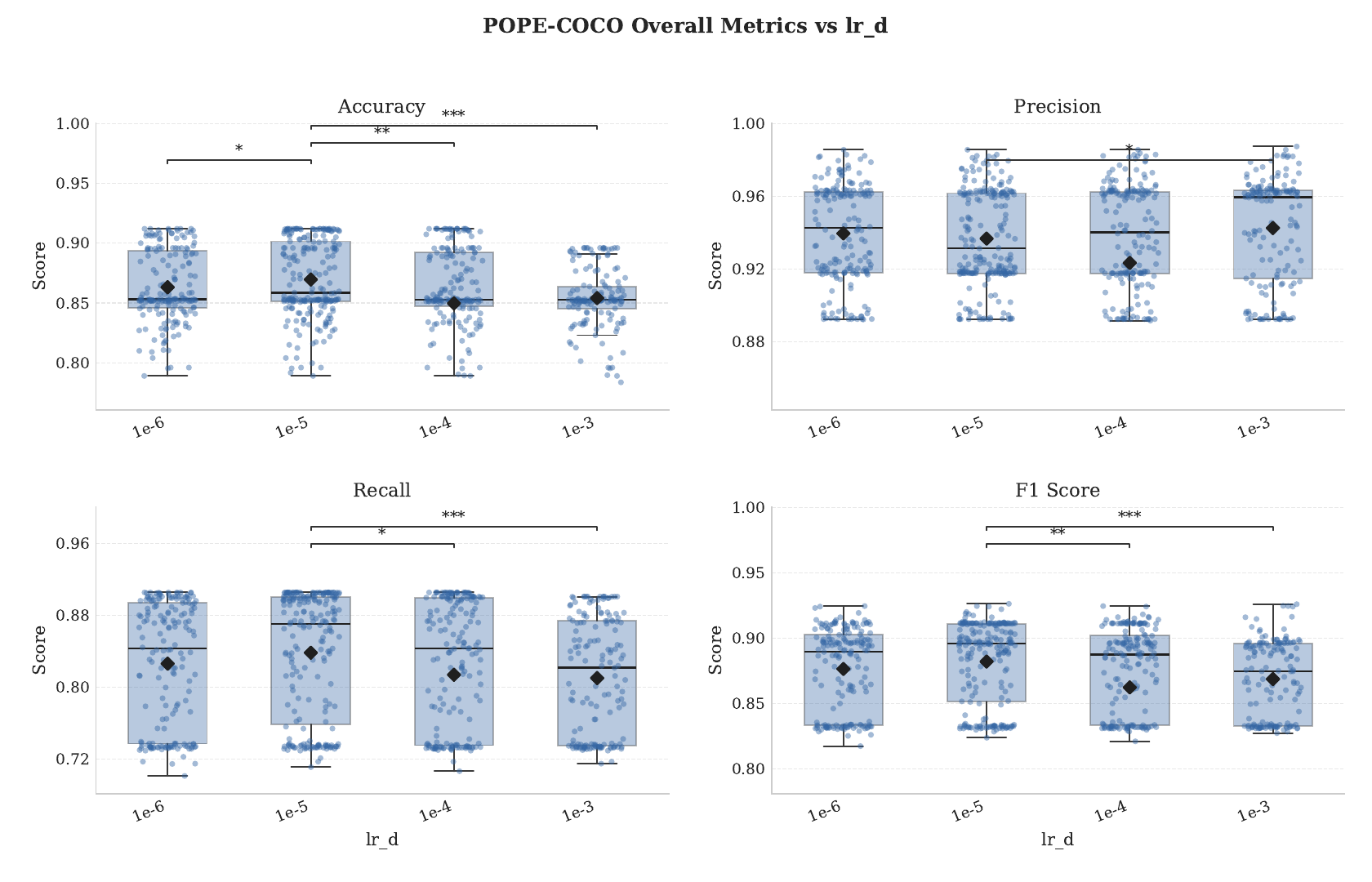}
\caption{POPE-COCO metrics vs.\ lr$_D$.}
\label{fig:supp_lr_d}
\end{figure}

\paragraph{Discriminator learning rate (lr$_D$).} \cref{fig:supp_lr_d} shows that smaller lr$_D$ (1e-5 to 1e-6) yields better Recall and F1, consistent with keeping the discriminator nearly frozen as a stable supervisory signal.

\begin{figure}[h]
\centering
\includegraphics[width=\columnwidth]{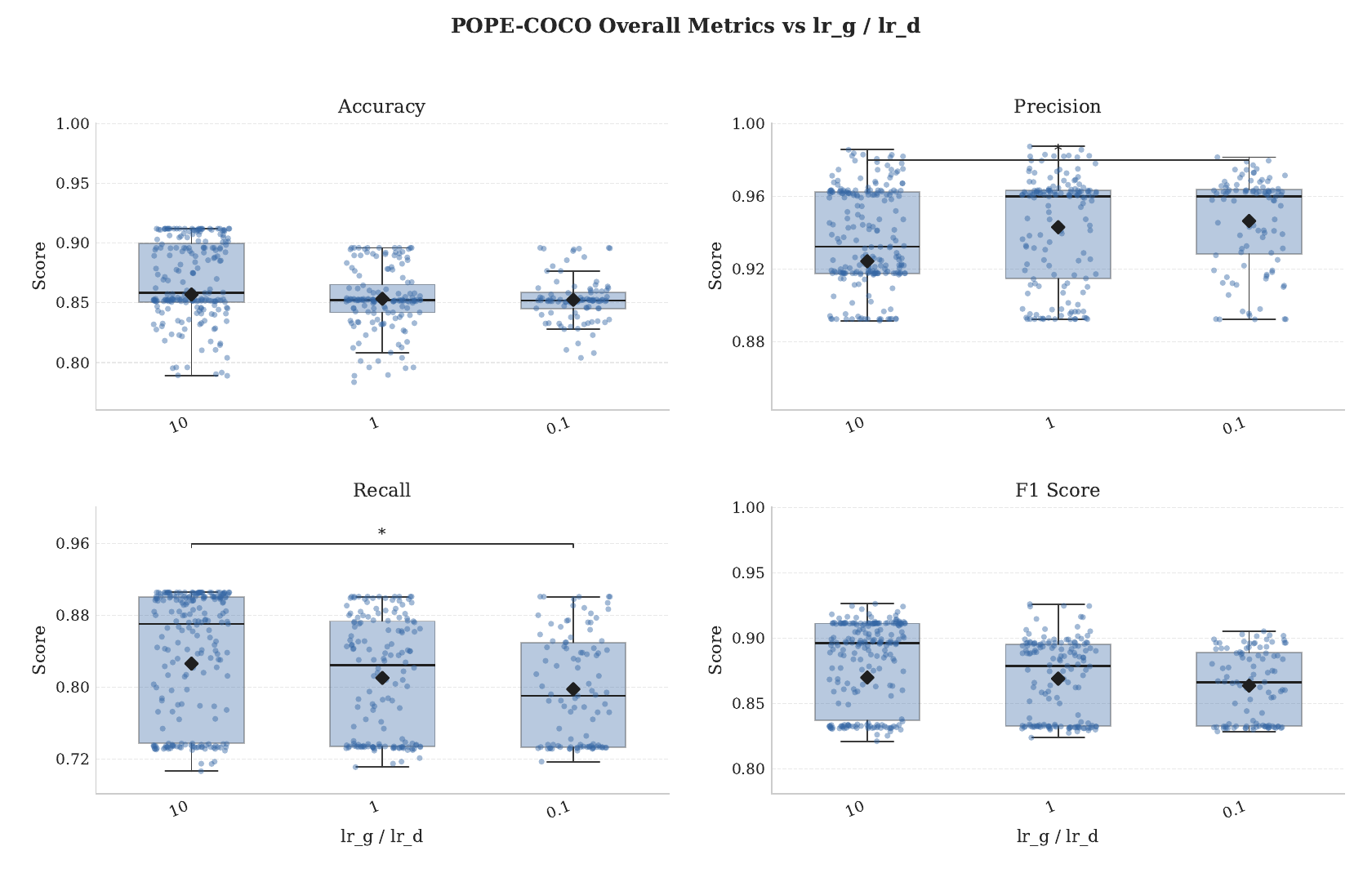}
\caption{POPE-COCO metrics vs.\ lr$_G$/lr$_D$ ratio.}
\label{fig:supp_lr_ratio}
\end{figure}

\paragraph{Learning rate ratio (lr$_G$/lr$_D$).} \cref{fig:supp_lr_ratio} shows that the ratio has limited impact on F1, suggesting MHSA is robust to the relative learning rate scale.

\begin{figure}[h]
\centering
\includegraphics[width=\columnwidth]{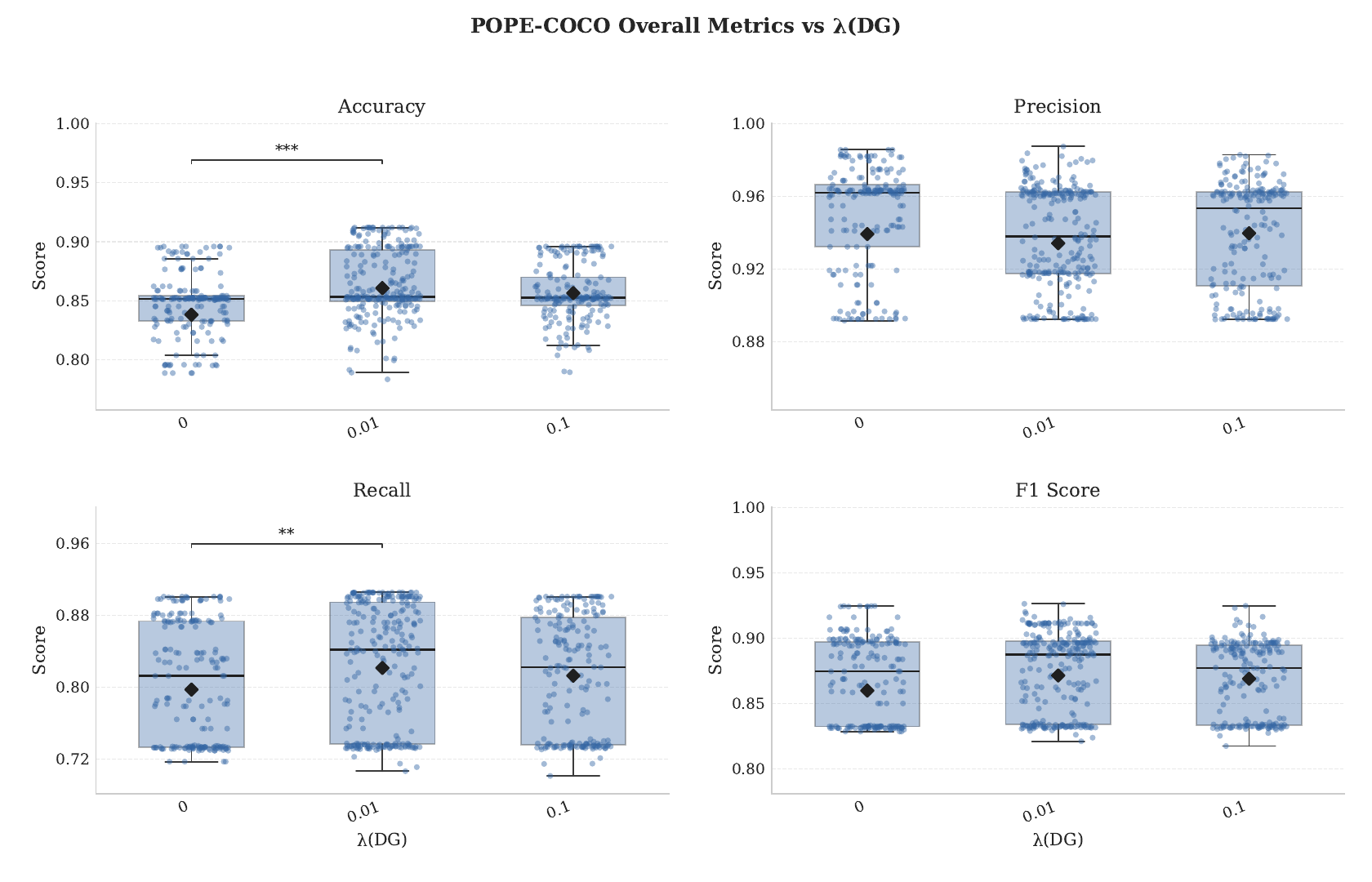}
\caption{POPE-COCO metrics vs.\ $\lambda_{\text{dg}}$.}
\label{fig:supp_alpha_adv}
\end{figure}

\paragraph{Detector-guided loss weight ($\lambda_{\text{dg}}$).} \cref{fig:supp_alpha_adv} shows that $\lambda_{\text{dg}}$=0.01 provides the best Accuracy. Removing it ($\lambda_{\text{dg}}$=0) degrades performance significantly.

\begin{figure}[h]
\centering
\includegraphics[width=\columnwidth]{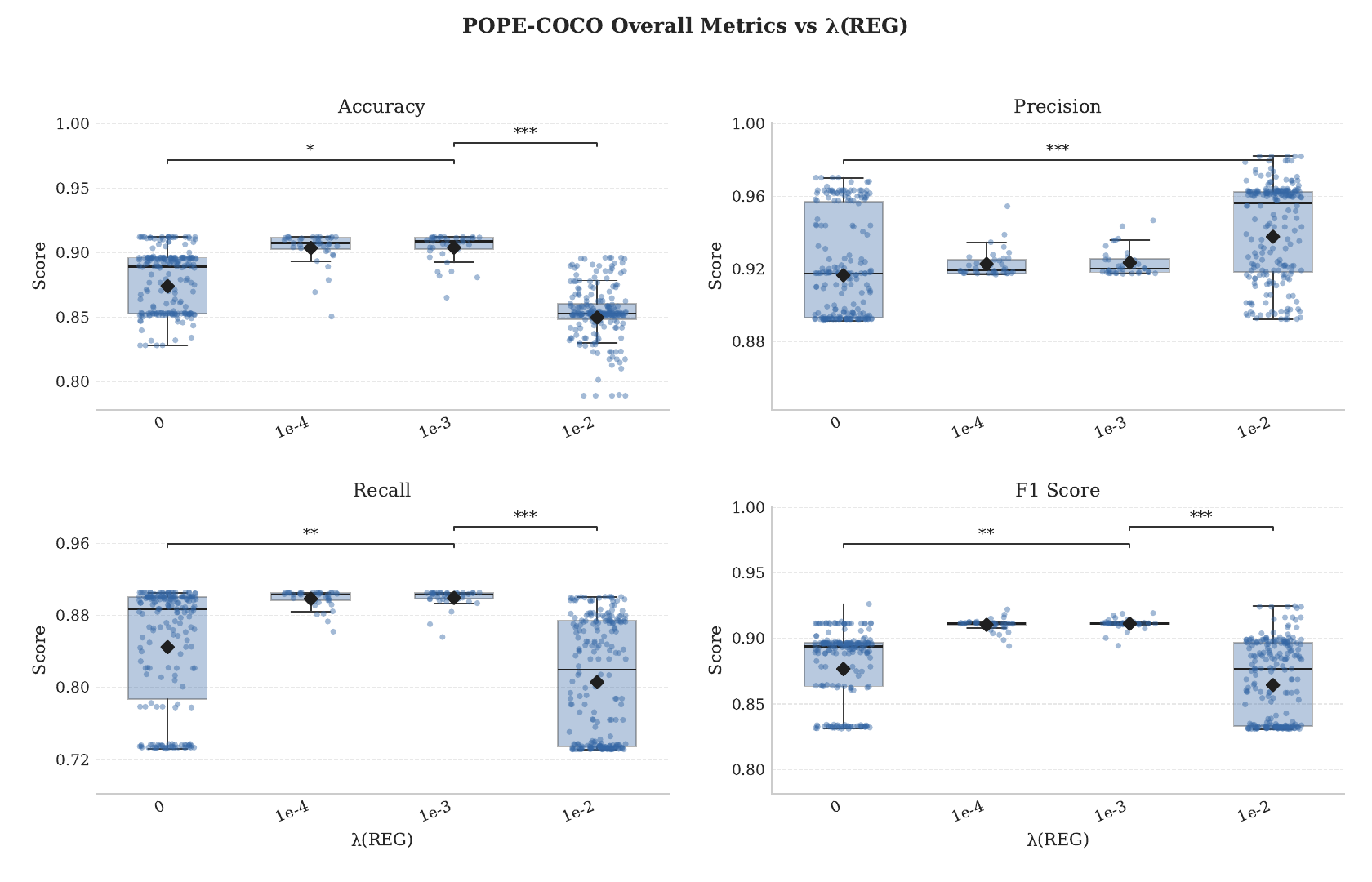}
\caption{POPE-COCO metrics vs.\ $\lambda_{\text{reg}}$.}
\label{fig:supp_reg_lambda}
\end{figure}

\paragraph{Regularization weight ($\lambda_{\text{reg}}$).} \cref{fig:supp_reg_lambda} shows that moderate regularization (1e-4 to 1e-3) achieves the best F1. Without regularization ($\lambda_{\text{reg}}$=0), variance increases; too strong (1e-2) suppresses useful corrections.

\begin{figure}[h]
\centering
\includegraphics[width=\columnwidth]{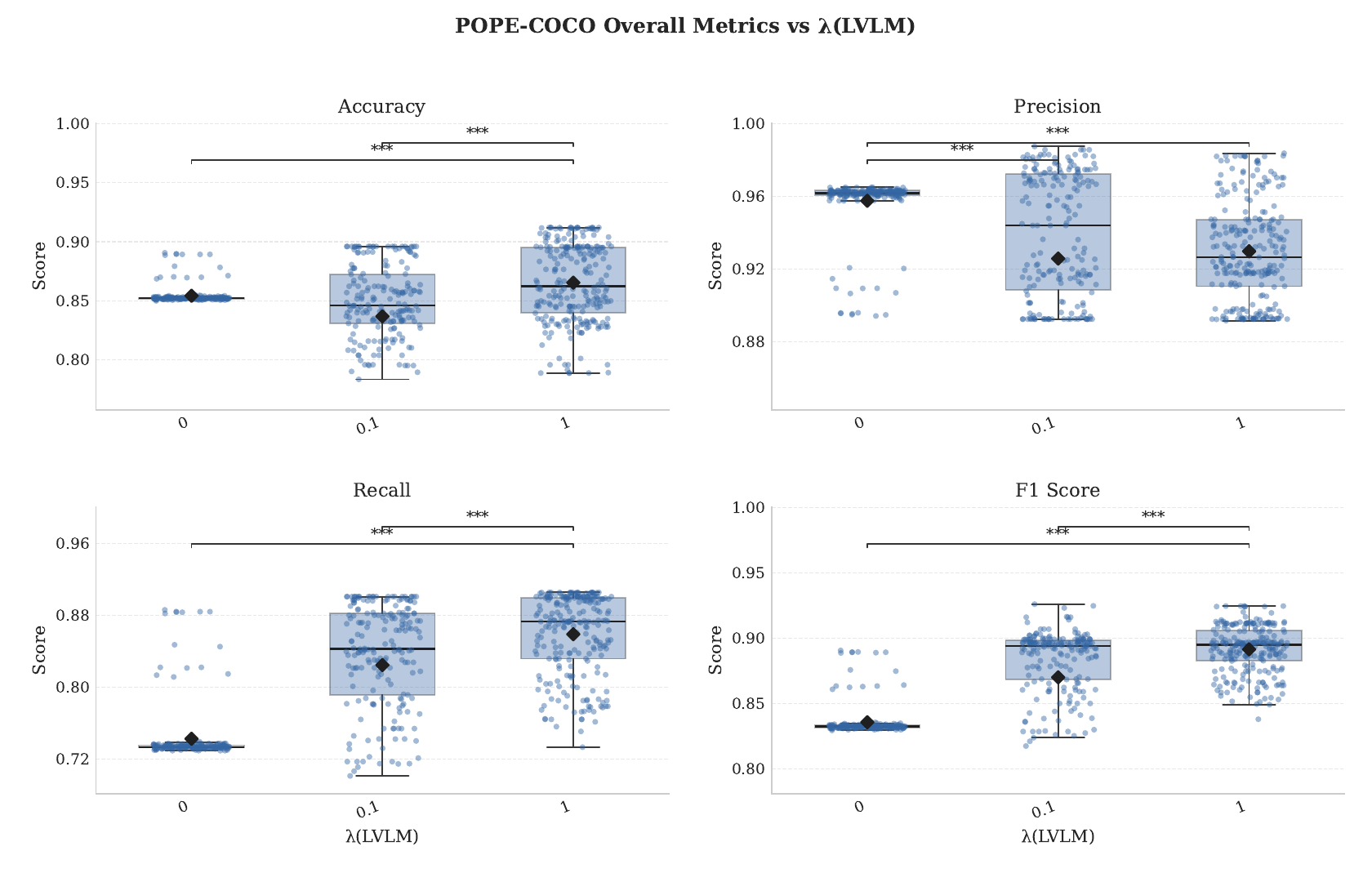}
\caption{POPE-COCO metrics vs.\ $\lambda_{\text{LVLM}}$.}
\label{fig:supp_alpha_llm}
\end{figure}

\paragraph{LVLM quality loss weight ($\lambda_{\text{LVLM}}$).} \cref{fig:supp_alpha_llm} shows that including $\mathcal{L}_{\text{LVLM}}$ ($\lambda_{\text{LVLM}} \geq 0.1$) substantially improves Recall and F1 over the no-LVLM-loss setting ($\lambda_{\text{LVLM}}$=0), validating its role in preserving output quality.

\section{Per-Category POPE Results}
\label{sec:appendix_percategory}

Per-category (Popular, Random, Adversarial) POPE results on MSCOCO, Objects365, and OpenImages (1,000 questions each). Best per metric in \textbf{bold} (including baseline).

\begin{table}[!htbp]
\centering
\caption{Per-category POPE: Qwen2.5-VL-7B on MSCOCO.}
\label{tab:appendix_qwen_coco}
\resizebox{\columnwidth}{!}{
\begin{tabular}{c|c|cccc}
    \toprule
    Category & Method & Accuracy & Precision & Recall & F1 \\
    \midrule
    \multirow{2}{*}{Popular} & Baseline & 85.90 & \textbf{95.54} & 76.94 & 85.24 \\
    & MHSA & \textbf{93.70} & 95.16 & \textbf{93.52} & \textbf{94.33} \\
    \midrule
    \multirow{2}{*}{Random} & Baseline & 88.20 & \textbf{98.40} & 76.72 & 86.21 \\
    & MHSA & \textbf{93.70} & 94.09 & \textbf{92.92} & \textbf{93.50} \\
    \midrule
    \multirow{2}{*}{Adversarial} & Baseline & 86.40 & \textbf{92.25} & 79.23 & 85.25 \\
    & MHSA & \textbf{90.90} & 88.57 & \textbf{93.75} & \textbf{91.09} \\
    \bottomrule
\end{tabular}
}
\end{table}

\begin{table}[!htbp]
\centering
\caption{Per-category POPE: Qwen2.5-VL-7B on Objects365.}
\label{tab:appendix_qwen_obj}
\resizebox{\columnwidth}{!}{
\begin{tabular}{c|c|cccc}
    \toprule
    Category & Method & Accuracy & Precision & Recall & F1 \\
    \midrule
    \multirow{2}{*}{Popular} & Baseline & 84.10 & 92.19 & 73.29 & 81.66 \\
    & MHSA & \textbf{92.70} & \textbf{93.25} & \textbf{91.51} & \textbf{92.37} \\
    \midrule
    \multirow{2}{*}{Random} & Baseline & 84.10 & \textbf{97.37} & 71.29 & 82.31 \\
    & MHSA & \textbf{91.90} & 94.69 & \textbf{89.40} & \textbf{91.97} \\
    \midrule
    \multirow{2}{*}{Adversarial} & Baseline & 83.60 & \textbf{89.36} & 76.06 & 82.17 \\
    & MHSA & \textbf{89.10} & 87.89 & \textbf{90.54} & \textbf{89.20} \\
    \bottomrule
\end{tabular}
}
\end{table}
\begin{table}[!htbp]
\centering
\caption{Per-category POPE: Qwen2.5-VL-7B on OpenImages.}
\label{tab:appendix_qwen_oimg}
\resizebox{\columnwidth}{!}{
\begin{tabular}{c|c|cccc}
    \toprule
    Category & Method & Accuracy & Precision & Recall & F1 \\
    \midrule
    \multirow{2}{*}{Popular} & Baseline & 83.20 & \textbf{78.97} & 88.80 & 83.59 \\
    & MHSA & \textbf{83.70} & 78.02 & \textbf{98.92} & \textbf{87.24} \\
    \midrule
    \multirow{2}{*}{Random} & Baseline & 93.20 & \textbf{98.20} & 87.93 & 92.78 \\
    & MHSA & \textbf{94.50} & 95.49 & \textbf{97.49} & \textbf{96.48} \\
    \midrule
    \multirow{2}{*}{Adversarial} & Baseline & \textbf{68.90} & \textbf{63.53} & 88.15 & 73.84 \\
    & MHSA & 65.30 & 61.13 & \textbf{98.96} & \textbf{75.57} \\
    \bottomrule
\end{tabular}
}
\end{table}

\begin{table}[!htbp]
\centering
\caption{Per-category POPE: InternVL2-8B on MSCOCO.}
\label{tab:appendix_intern_coco}
\resizebox{\columnwidth}{!}{
\begin{tabular}{c|c|cccc}
    \toprule
    Category & Method & Accuracy & Precision & Recall & F1 \\
    \midrule
    \multirow{2}{*}{Popular} & Baseline & 86.50 & 92.64 & 80.91 & 86.38 \\
    & MHSA & \textbf{94.90} & \textbf{92.68} & \textbf{98.11} & \textbf{95.32} \\
    \midrule
    \multirow{2}{*}{Random} & Baseline & 89.70 & \textbf{94.16} & 83.78 & 88.67 \\
    & MHSA & \textbf{95.70} & 92.44 & \textbf{99.17} & \textbf{95.69} \\
    \midrule
    \multirow{2}{*}{Adversarial} & Baseline & 85.00 & \textbf{86.19} & 83.06 & 84.60 \\
    & MHSA & \textbf{91.00} & 85.61 & \textbf{98.39} & \textbf{91.56} \\
    \bottomrule
\end{tabular}
}
\end{table}

\begin{table}[!htbp]
\centering
\caption{Per-category POPE: InternVL2-8B on Objects365.}
\label{tab:appendix_intern_obj}
\resizebox{\columnwidth}{!}{
\begin{tabular}{c|c|cccc}
    \toprule
    Category & Method & Accuracy & Precision & Recall & F1 \\
    \midrule
    \multirow{2}{*}{Popular} & Baseline & 85.80 & 92.16 & 78.07 & 84.53 \\
    & MHSA & \textbf{94.10} & \textbf{94.69} & \textbf{93.36} & \textbf{94.02} \\
    \midrule
    \multirow{2}{*}{Random} & Baseline & 85.60 & \textbf{92.24} & 77.93 & 84.48 \\
    & MHSA & \textbf{88.70} & 87.50 & \textbf{90.46} & \textbf{88.95} \\
    \midrule
    \multirow{2}{*}{Adversarial} & Baseline & 80.50 & 84.65 & 74.70 & 79.37 \\
    & MHSA & \textbf{88.80} & \textbf{85.85} & \textbf{93.03} & \textbf{89.29} \\
    \bottomrule
\end{tabular}
}
\end{table}

\begin{table}[!htbp]
\centering
\caption{Per-category POPE: InternVL2-8B on OpenImages.}
\label{tab:appendix_intern_oimg}
\resizebox{\columnwidth}{!}{
\begin{tabular}{c|c|cccc}
    \toprule
    Category & Method & Accuracy & Precision & Recall & F1 \\
    \midrule
    \multirow{2}{*}{Popular} & Baseline & 80.70 & 75.39 & 89.51 & 81.84 \\
    & MHSA & \textbf{86.00} & \textbf{77.72} & \textbf{99.79} & \textbf{87.39} \\
    \midrule
    \multirow{2}{*}{Random} & Baseline & 92.80 & \textbf{94.61} & 91.33 & 92.94 \\
    & MHSA & \textbf{95.80} & 92.67 & \textbf{99.81} & \textbf{96.10} \\
    \midrule
    \multirow{2}{*}{Adversarial} & Baseline & 65.10 & 60.86 & 88.43 & 72.10 \\
    & MHSA & \textbf{69.40} & \textbf{62.56} & \textbf{99.61} & \textbf{76.85} \\
    \bottomrule
\end{tabular}
}
\end{table}

\begin{table}[!htbp]
\centering
\caption{Per-category POPE: LLaVA-v1.5-7B on MSCOCO.}
\label{tab:appendix_llava_coco}
\resizebox{\columnwidth}{!}{
\begin{tabular}{c|c|cccc}
    \toprule
    Category & Method & Accuracy & Precision & Recall & F1 \\
    \midrule
    \multirow{2}{*}{Popular} & Baseline & 83.80 & \textbf{91.80} & 76.18 & 83.26 \\
    & MHSA & \textbf{91.90} & 91.33 & \textbf{93.57} & \textbf{92.44} \\
    \midrule
    \multirow{2}{*}{Random} & Baseline & 88.60 & \textbf{96.69} & 79.00 & 86.96 \\
    & MHSA & \textbf{94.80} & 95.35 & \textbf{93.76} & \textbf{94.55} \\
    \midrule
    \multirow{2}{*}{Adversarial} & Baseline & 84.30 & \textbf{86.77} & 80.65 & 83.59 \\
    & MHSA & \textbf{89.60} & 86.03 & \textbf{94.35} & \textbf{90.00} \\
    \bottomrule
\end{tabular}
}
\end{table}

\begin{table}[!htbp]
\centering
\caption{Per-category POPE: LLaVA-v1.5-7B on Objects365.}
\label{tab:appendix_llava_obj}
\resizebox{\columnwidth}{!}{
\begin{tabular}{c|c|cccc}
    \toprule
    Category & Method & Accuracy & Precision & Recall & F1 \\
    \midrule
    \multirow{2}{*}{Popular} & Baseline & 84.30 & \textbf{94.27} & 72.84 & 82.18 \\
    & MHSA & \textbf{93.00} & 90.40 & \textbf{97.56} & \textbf{93.84} \\
    \midrule
    \multirow{2}{*}{Random} & Baseline & 85.30 & \textbf{94.28} & 75.35 & 83.76 \\
    & MHSA & \textbf{91.00} & 86.64 & \textbf{98.21} & \textbf{92.06} \\
    \midrule
    \multirow{2}{*}{Adversarial} & Baseline & 78.10 & \textbf{83.61} & 70.12 & 76.27 \\
    & MHSA & \textbf{87.10} & 80.62 & \textbf{98.59} & \textbf{88.71} \\
    \bottomrule
\end{tabular}
}
\end{table}
\begin{table}[!htbp]
\centering
\caption{Per-category POPE: LLaVA-v1.5-7B on OpenImages.}
\label{tab:appendix_llava_oimg}
\resizebox{\columnwidth}{!}{
\begin{tabular}{c|c|cccc}
    \toprule
    Category & Method & Accuracy & Precision & Recall & F1 \\
    \midrule
    \multirow{2}{*}{Popular} & Baseline & 79.90 & 73.94 & 90.29 & 81.30 \\
    & MHSA & \textbf{82.40} & \textbf{75.08} & \textbf{96.27} & \textbf{84.36} \\
    \midrule
    \multirow{2}{*}{Random} & Baseline & 92.70 & \textbf{94.97} & 89.67 & 92.24 \\
    & MHSA & \textbf{94.60} & 93.72 & \textbf{95.86} & \textbf{94.78} \\
    \midrule
    \multirow{2}{*}{Adversarial} & Baseline & 62.90 & 58.66 & 90.32 & 71.13 \\
    & MHSA & \textbf{66.20} & \textbf{61.25} & \textbf{96.02} & \textbf{74.79} \\
    \bottomrule
\end{tabular}
}
\end{table}

\section{Cross-Dataset OOD Generalization: Full Metrics}
\label{sec:appendix_cross}

To evaluate whether MHSA's learned corrections generalize beyond the training distribution, we conduct cross-dataset experiments: training on POPE-COCO and testing on POPE-Objects365, and vice versa. \cref{tab:appendix_cross_qwen,tab:appendix_cross_intern,tab:appendix_cross_llava} report per-category (Popular, Random, Adversarial) results for all four metrics. For each metric, ``Train'' indicates which dataset MHSA was trained on; best results per test column (excluding the baseline) are in \textbf{bold}.

%%% === Qwen ===
\begin{table}[!htbp]
\centering
\caption{Cross-dataset generalization: Qwen2.5-VL-7B.}
\label{tab:appendix_cross_qwen}
\resizebox{\columnwidth}{!}{
\begin{tabular}{c|c|c|cc}
    \toprule
    & & & \multicolumn{2}{c}{Test} \\
    \cmidrule(lr){4-5}
    Category & Metric & Train & COCO & Obj365 \\
    \midrule
    \multirow{12}{*}{Popular}
    & \multirow{3}{*}{Accuracy} & COCO & 93.70 & 92.20 \\
    & & Obj365 & \textbf{94.00} & \textbf{92.70} \\
    & & Baseline & 85.90 & 84.10 \\
    \cmidrule{2-5}
    & \multirow{3}{*}{Precision} & COCO & 95.16 & \textbf{94.93} \\
    & & Obj365 & 94.16 & 93.25 \\
    & & Baseline & \textbf{95.54} & 92.19 \\
    \cmidrule{2-5}
    & \multirow{3}{*}{Recall} & COCO & 93.52 & 90.17 \\
    & & Obj365 & \textbf{94.52} & \textbf{91.51} \\
    & & Baseline & 76.94 & 73.29 \\
    \cmidrule{2-5}
    & \multirow{3}{*}{F1} & COCO & 94.33 & \textbf{92.49} \\
    & & Obj365 & \textbf{94.34} & 92.37 \\
    & & Baseline & 85.24 & 81.66 \\
    \midrule
    \multirow{12}{*}{Random}
    & \multirow{3}{*}{Accuracy} & COCO & 93.70 & 90.90 \\
    & & Obj365 & \textbf{94.90} & \textbf{91.90} \\
    & & Baseline & 88.20 & 84.10 \\
    \cmidrule{2-5}
    & \multirow{3}{*}{Precision} & COCO & 94.09 & 94.97 \\
    & & Obj365 & 94.06 & 94.69 \\
    & & Baseline & \textbf{98.40} & \textbf{97.37} \\
    \cmidrule{2-5}
    & \multirow{3}{*}{Recall} & COCO & 92.92 & 88.30 \\
    & & Obj365 & \textbf{95.43} & \textbf{89.40} \\
    & & Baseline & 76.72 & 71.29 \\
    \cmidrule{2-5}
    & \multirow{3}{*}{F1} & COCO & 93.50 & 91.52 \\
    & & Obj365 & \textbf{94.74} & \textbf{91.97} \\
    & & Baseline & 86.21 & 82.31 \\
    \midrule
    \multirow{12}{*}{Adversarial}
    & \multirow{3}{*}{Accuracy} & COCO & 90.90 & 88.80 \\
    & & Obj365 & \textbf{91.10} & \textbf{89.10} \\
    & & Baseline & 86.40 & 83.60 \\
    \cmidrule{2-5}
    & \multirow{3}{*}{Precision} & COCO & 88.57 & 88.62 \\
    & & Obj365 & 87.48 & 87.89 \\
    & & Baseline & \textbf{92.25} & \textbf{89.36} \\
    \cmidrule{2-5}
    & \multirow{3}{*}{Recall} & COCO & 93.75 & 89.70 \\
    & & Obj365 & \textbf{95.77} & \textbf{90.54} \\
    & & Baseline & 79.23 & 76.06 \\
    \cmidrule{2-5}
    & \multirow{3}{*}{F1} & COCO & 91.09 & 89.16 \\
    & & Obj365 & \textbf{91.43} & \textbf{89.20} \\
    & & Baseline & 85.25 & 82.17 \\
    \bottomrule
\end{tabular}
}
\end{table}

%%% === InternVL ===
\begin{table}[!htbp]
\centering
\caption{Cross-dataset generalization: InternVL2-8B.}
\label{tab:appendix_cross_intern}
\resizebox{\columnwidth}{!}{
\begin{tabular}{c|c|c|cc}
    \toprule
    & & & \multicolumn{2}{c}{Test} \\
    \cmidrule(lr){4-5}
    Category & Metric & Train & COCO & Obj365 \\
    \midrule
    \multirow{12}{*}{Popular}
    & \multirow{3}{*}{Accuracy} & COCO & 94.90 & 92.70 \\
    & & Obj365 & \textbf{97.20} & \textbf{94.10} \\
    & & Baseline & 86.50 & 85.80 \\
    \cmidrule{2-5}
    & \multirow{3}{*}{Precision} & COCO & 92.68 & 88.83 \\
    & & Obj365 & \textbf{97.53} & \textbf{94.69} \\
    & & Baseline & 92.64 & 92.16 \\
    \cmidrule{2-5}
    & \multirow{3}{*}{Recall} & COCO & \textbf{98.11} & \textbf{97.59} \\
    & & Obj365 & 97.16 & 93.36 \\
    & & Baseline & 80.91 & 78.07 \\
    \cmidrule{2-5}
    & \multirow{3}{*}{F1} & COCO & 95.32 & 93.00 \\
    & & Obj365 & \textbf{97.35} & \textbf{94.02} \\
    & & Baseline & 86.38 & 84.53 \\
    \midrule
    \multirow{12}{*}{Random}
    & \multirow{3}{*}{Accuracy} & COCO & 95.70 & \textbf{90.00} \\
    & & Obj365 & \textbf{96.00} & 88.70 \\
    & & Baseline & 89.70 & 85.60 \\
    \cmidrule{2-5}
    & \multirow{3}{*}{Precision} & COCO & 92.44 & 85.04 \\
    & & Obj365 & \textbf{95.09} & 87.50 \\
    & & Baseline & 94.16 & \textbf{92.24} \\
    \cmidrule{2-5}
    & \multirow{3}{*}{Recall} & COCO & \textbf{99.17} & \textbf{97.22} \\
    & & Obj365 & 96.67 & 90.46 \\
    & & Baseline & 83.78 & 77.93 \\
    \cmidrule{2-5}
    & \multirow{3}{*}{F1} & COCO & 95.69 & \textbf{90.72} \\
    & & Obj365 & \textbf{95.88} & 88.95 \\
    & & Baseline & 88.67 & 84.48 \\
    \midrule
    \multirow{12}{*}{Adversarial}
    & \multirow{3}{*}{Accuracy} & COCO & 91.00 & 86.80 \\
    & & Obj365 & \textbf{93.90} & \textbf{88.80} \\
    & & Baseline & 85.00 & 80.50 \\
    \cmidrule{2-5}
    & \multirow{3}{*}{Precision} & COCO & 85.61 & 80.33 \\
    & & Obj365 & \textbf{90.96} & \textbf{85.85} \\
    & & Baseline & 86.19 & 84.65 \\
    \cmidrule{2-5}
    & \multirow{3}{*}{Recall} & COCO & \textbf{98.39} & \textbf{97.61} \\
    & & Obj365 & 97.38 & 93.03 \\
    & & Baseline & 83.06 & 74.70 \\
    \cmidrule{2-5}
    & \multirow{3}{*}{F1} & COCO & 91.56 & 88.13 \\
    & & Obj365 & \textbf{94.06} & \textbf{89.29} \\
    & & Baseline & 84.60 & 79.37 \\
    \bottomrule
\end{tabular}
}
\end{table}

%%% === LLaVA ===
\begin{table}[!htbp]
\centering
\caption{Cross-dataset generalization: LLaVA-v1.5-7B.}
\label{tab:appendix_cross_llava}
\resizebox{\columnwidth}{!}{
\begin{tabular}{c|c|c|cc}
    \toprule
    & & & \multicolumn{2}{c}{Test} \\
    \cmidrule(lr){4-5}
    Category & Metric & Train & COCO & Obj365 \\
    \midrule
    \multirow{12}{*}{Popular}
    & \multirow{3}{*}{Accuracy} & COCO & 91.90 & 91.40 \\
    & & Obj365 & \textbf{93.50} & \textbf{93.00} \\
    & & Baseline & 83.80 & 84.30 \\
    \cmidrule{2-5}
    & \multirow{3}{*}{Precision} & COCO & 91.33 & 91.02 \\
    & & Obj365 & 91.58 & 90.40 \\
    & & Baseline & \textbf{91.80} & \textbf{94.27} \\
    \cmidrule{2-5}
    & \multirow{3}{*}{Recall} & COCO & 93.57 & 91.75 \\
    & & Obj365 & \textbf{98.08} & \textbf{97.56} \\
    & & Baseline & 76.18 & 72.84 \\
    \cmidrule{2-5}
    & \multirow{3}{*}{F1} & COCO & 92.44 & 91.38 \\
    & & Obj365 & \textbf{94.72} & \textbf{93.84} \\
    & & Baseline & 83.26 & 82.18 \\
    \midrule
    \multirow{12}{*}{Random}
    & \multirow{3}{*}{Accuracy} & COCO & 94.80 & 89.80 \\
    & & Obj365 & \textbf{95.70} & \textbf{91.00} \\
    & & Baseline & 88.60 & 85.30 \\
    \cmidrule{2-5}
    & \multirow{3}{*}{Precision} & COCO & 95.35 & 87.62 \\
    & & Obj365 & 95.08 & 86.64 \\
    & & Baseline & \textbf{96.69} & \textbf{94.28} \\
    \cmidrule{2-5}
    & \multirow{3}{*}{Recall} & COCO & 93.76 & 92.84 \\
    & & Obj365 & \textbf{97.27} & \textbf{98.21} \\
    & & Baseline & 79.00 & 75.35 \\
    \cmidrule{2-5}
    & \multirow{3}{*}{F1} & COCO & 94.55 & 90.15 \\
    & & Obj365 & \textbf{96.17} & \textbf{92.06} \\
    & & Baseline & 86.96 & 83.76 \\
    \midrule
    \multirow{12}{*}{Adversarial}
    & \multirow{3}{*}{Accuracy} & COCO & 89.60 & 86.20 \\
    & & Obj365 & \textbf{90.30} & \textbf{87.10} \\
    & & Baseline & 84.30 & 78.10 \\
    \cmidrule{2-5}
    & \multirow{3}{*}{Precision} & COCO & 86.03 & 81.93 \\
    & & Obj365 & 85.59 & 80.62 \\
    & & Baseline & \textbf{86.77} & \textbf{83.61} \\
    \cmidrule{2-5}
    & \multirow{3}{*}{Recall} & COCO & 94.35 & 93.03 \\
    & & Obj365 & \textbf{97.57} & \textbf{98.59} \\
    & & Baseline & 80.65 & 70.12 \\
    \cmidrule{2-5}
    & \multirow{3}{*}{F1} & COCO & 90.00 & 87.13 \\
    & & Obj365 & \textbf{91.18} & \textbf{88.71} \\
    & & Baseline & 83.59 & 76.27 \\
    \bottomrule
\end{tabular}
}
\end{table}
As shown in \cref{tab:appendix_cross_qwen}, MHSA trained on either dataset consistently outperforms the baseline on Qwen2.5-VL-7B across all categories and metrics. The Obj365-trained model slightly edges out the COCO-trained one in most Accuracy and Recall comparisons, while the COCO-trained model tends to retain higher Precision. Notably, the cross-dataset gap is small (typically $<$1\% F1), indicating that the learned attention corrections transfer well.

\cref{tab:appendix_cross_intern} shows a similar pattern for InternVL2-8B. The Obj365-trained model achieves the highest F1 in most settings, with particularly strong gains on the Adversarial category (up to +9.4\% F1 over baseline on COCO). The COCO-trained model, on the other hand, achieves the highest Recall across all categories, suggesting a more aggressive correction strategy. Both cross-dataset configurations substantially outperform the baseline, confirming robust OOD transfer.

\cref{tab:appendix_cross_llava} confirms the same trend on LLaVA-v1.5-7B. Cross-dataset F1 improvements over the baseline range from +6.4\% to +12.4\%, with the largest gains appearing on the Adversarial category of Objects365. Across all three models, both in-distribution and out-of-distribution MHSA configurations consistently and substantially outperform the uncorrected baseline, demonstrating that the attention correction patterns learned by MHSA are not dataset-specific but capture general hallucination-related attention signatures.
% \clearpage
\section{Ablation: Generative Task (Caption)}
\label{sec:appendix_cap_ablation}

\cref{tab:appendix_cap_ablation} ablates the loss components for token-level MHSA on image captioning (Flickr30k, Qwen2.5-VL-7B, 1,000 images). The final configuration uses $\lambda_{\text{dg}}{=}0.5$, $\lambda_{\text{reg}}{=}0.01$, $\lambda_{\text{LVLM}}{=}0$. Each variant changes one term.

\begin{table}[!htbp]
\centering
\caption{Caption ablation (Flickr30k, Qwen2.5-VL-7B). CHAIR$_i$/CHAIR$_s$ $\downarrow$, Recall $\uparrow$. Best trade-off in \textbf{bold}.}
\label{tab:appendix_cap_ablation}
\resizebox{\columnwidth}{!}{
\begin{tabular}{l|ccc|ccc}
    \toprule
    Variant & CHAIR$_i$ $\downarrow$ & CHAIR$_s$ $\downarrow$ & Recall $\uparrow$ & $\Delta$CHAIR$_i$ $\downarrow$ & $\Delta$CHAIR$_s$ $\downarrow$ & $\Delta$Recall $\uparrow$ \\
    \midrule
    Baseline (no correction) & 16.43 & 37.50 & 86.63 & --- & --- & --- \\
    \midrule
    \textbf{Ours (full)} & \textbf{9.20} & \textbf{21.00} & \textbf{83.28} & $-$7.23 & $-$16.50 & $-$3.35 \\
    \midrule
    w/o $\mathcal{L}_{\text{dg}}$ & 16.74 & 37.50 & 86.63 & $+$0.31 & $\pm$0.00 & $\pm$0.00 \\
    w/o $\mathcal{L}_{\text{reg}}$ & 4.41 & 8.90 & 72.59 & $-$12.02 & $-$28.60 & $-$14.04 \\
    w/ $\mathcal{L}_{\text{LVLM}}$ & 3.42 & 6.40 & 71.47 & $-$13.01 & $-$31.10 & $-$15.16 \\
    \bottomrule
\end{tabular}
}
\end{table}
Without $\mathcal{L}_{\text{dg}}$, only $\mathcal{L}_{\text{reg}}$ remains, which pushes $\Delta\mathbf{A}$ toward zero and produces virtually no correction ($\Delta$CHAIR$_i{=}+0.31$, Recall unchanged). This confirms that $\mathcal{L}_{\text{dg}}$ is the essential driving force behind hallucination reduction. Removing $\mathcal{L}_{\text{reg}}$ or adding $\mathcal{L}_{\text{LVLM}}$ yields much lower CHAIR ($-$73\%/$-$79\%) but severe Recall drops ($-$14\%/$-$15\%), indicating aggressive generation suppression. The full configuration best balances hallucination reduction ($-$44\% CHAIR$_i$) with content preservation ($-$3.4\% Recall).

Note that unlike the discriminative (POPE) setting where $\mathcal{L}_{\text{LVLM}}$ supervises each sample via the ground-truth Yes/No label, the caption setting operates with ($\lambda_{\text{LVLM}}{=}0$, \cref{tab:appendix_hyperparams}): the generator trains on pre-extracted attention without back-propagating through the LVLM. Including $\mathcal{L}_{\text{LVLM}}$ here over-constrains the generator, causing it to suppress generation rather than make targeted corrections.

\section{Additional Attention Visualizations: Discriminative Task (POPE)}
\label{sec:appendix_attn_pope}
 
\begin{figure*}[!htbp]
\centering
\includegraphics[width=0.82\textwidth]{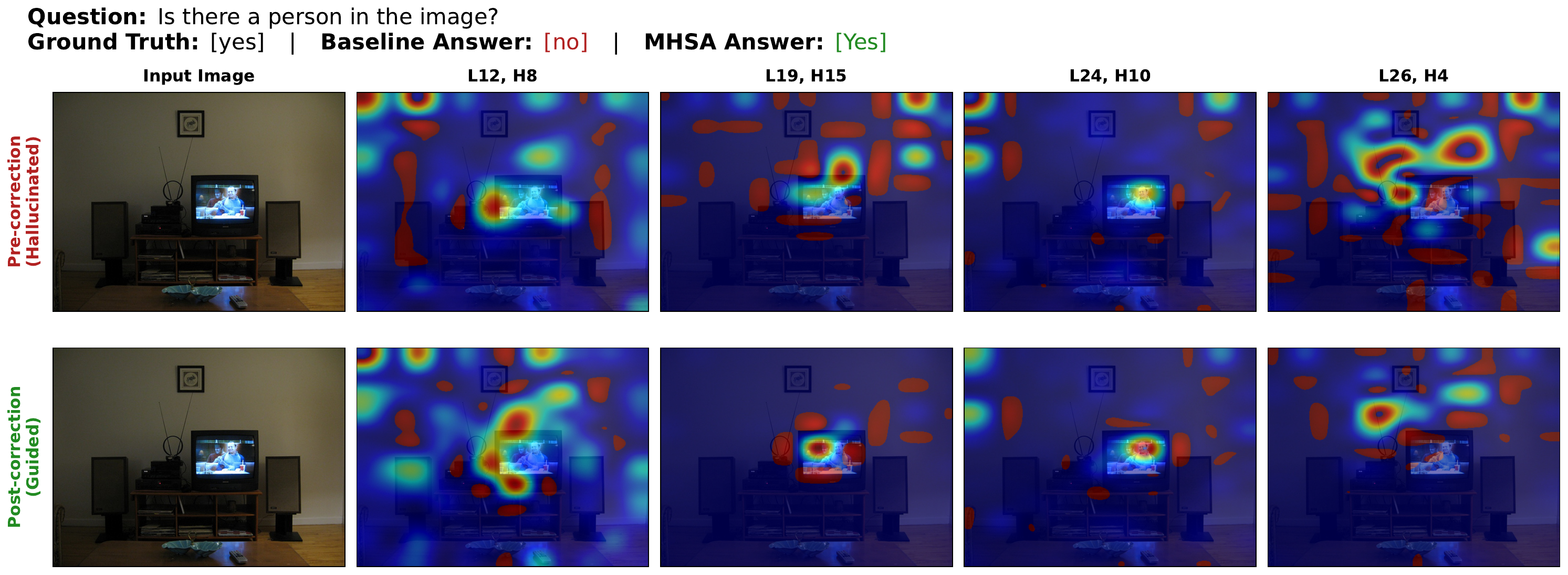}
\caption{POPE attention visualization for Qwen2.5-VL-7B. Question: ``Is there a person in the image?'' (GT: Yes). Pre-correction attention is spatially diffuse and fails to localize the central subject. Post-MHSA attention concentrates on the person region, correcting the prediction from No to Yes.}
\label{fig:supp_pope_qwen}
\end{figure*}

\begin{figure*}[!htbp]
\centering
\includegraphics[width=0.82\textwidth]{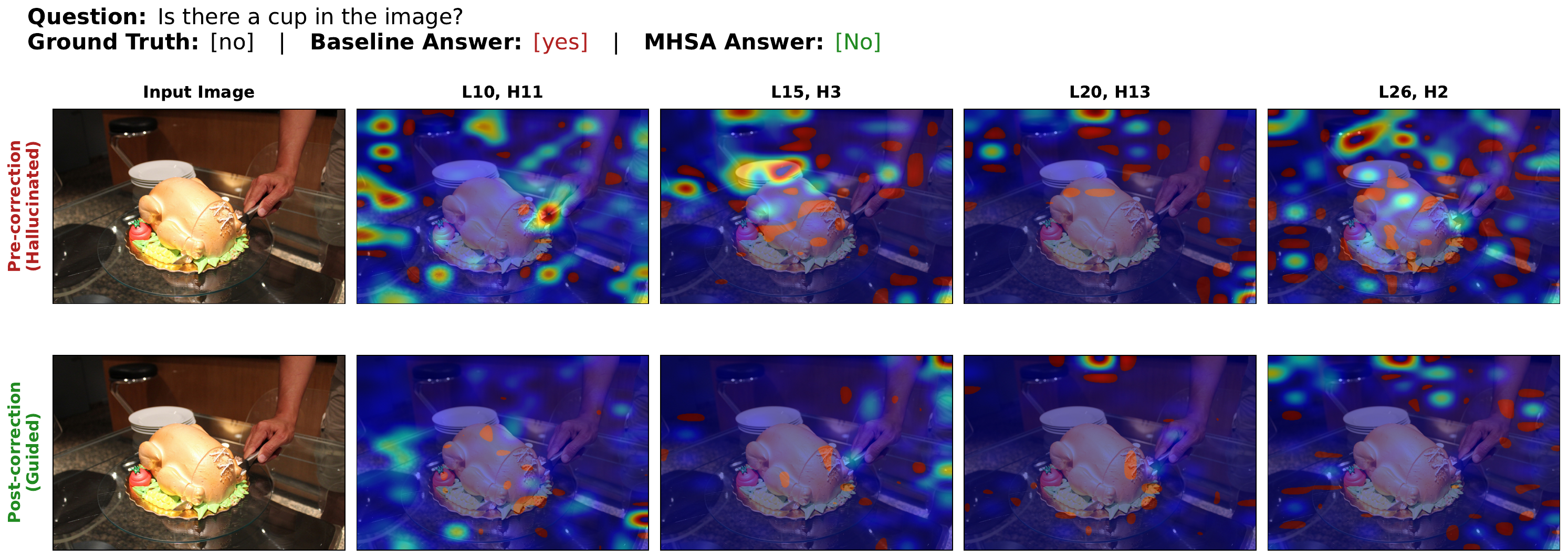}
\caption{POPE attention visualization for InternVL2-8B. Question: ``Is there a cup in the image?'' (GT: No). The baseline assigns noisy attention to background items, causing a false-positive ``Yes.'' MHSA suppresses these spurious activations, correcting the prediction to No.}
\label{fig:supp_pope_intern}
\end{figure*}

\begin{figure*}[!htbp]
\centering
\includegraphics[width=0.82\textwidth]{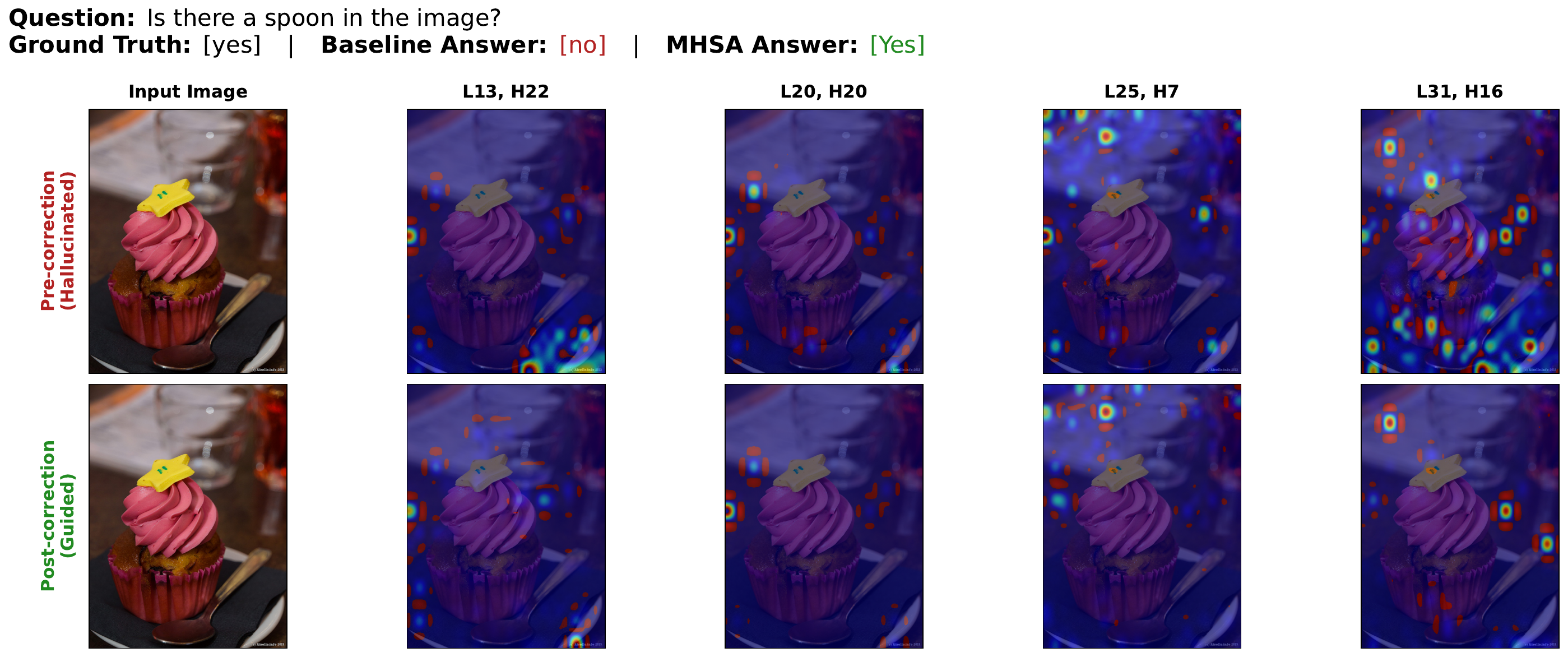}
\caption{POPE attention visualization for LLaVA-v1.5-7B. Question: ``Is there a spoon in the image?'' (GT: Yes). Baseline attention fails to localize the spoon, producing an incorrect ``No.'' Post-MHSA attention is redirected toward the spoon's location, correcting the prediction to Yes.}
\label{fig:supp_pope_llava}
\end{figure*}

To further validate the effectiveness and generalizability of MHSA, we provide additional qualitative visualizations across all three LVLMs, encompassing both false-negative hallucinations (missing present objects) and false-positive hallucinations (identifying absent objects). For Qwen2.5-VL-7B (\cref{fig:supp_pope_qwen}), the baseline provides a hallucinated ``No'' due to diffuse, unfocused attention, while MHSA concentrates attention on the person region and corrects the answer to ``Yes.'' For InternVL2-8B (\cref{fig:supp_pope_intern}), the baseline falsely answers ``Yes'' to a cup query; MHSA suppresses noisy background attention, yielding the correct ``No.'' For LLaVA-v1.5-7B (\cref{fig:supp_pope_llava}), the baseline fails to localize the spoon and predicts ``No,'' whereas MHSA redirects attention to the target object and corrects the answer to ``Yes.'' Across all three models, a consistent pattern emerges: pre-correction attention tends to be diffuse or misaligned with the queried object, while post-correction attention is spatially concentrated on the task-relevant region, consistent with the mechanistic analysis in Sec.~4.8 of the main paper.

\begin{figure*}[!htbp]
\centering
\includegraphics[width=0.93\textwidth]{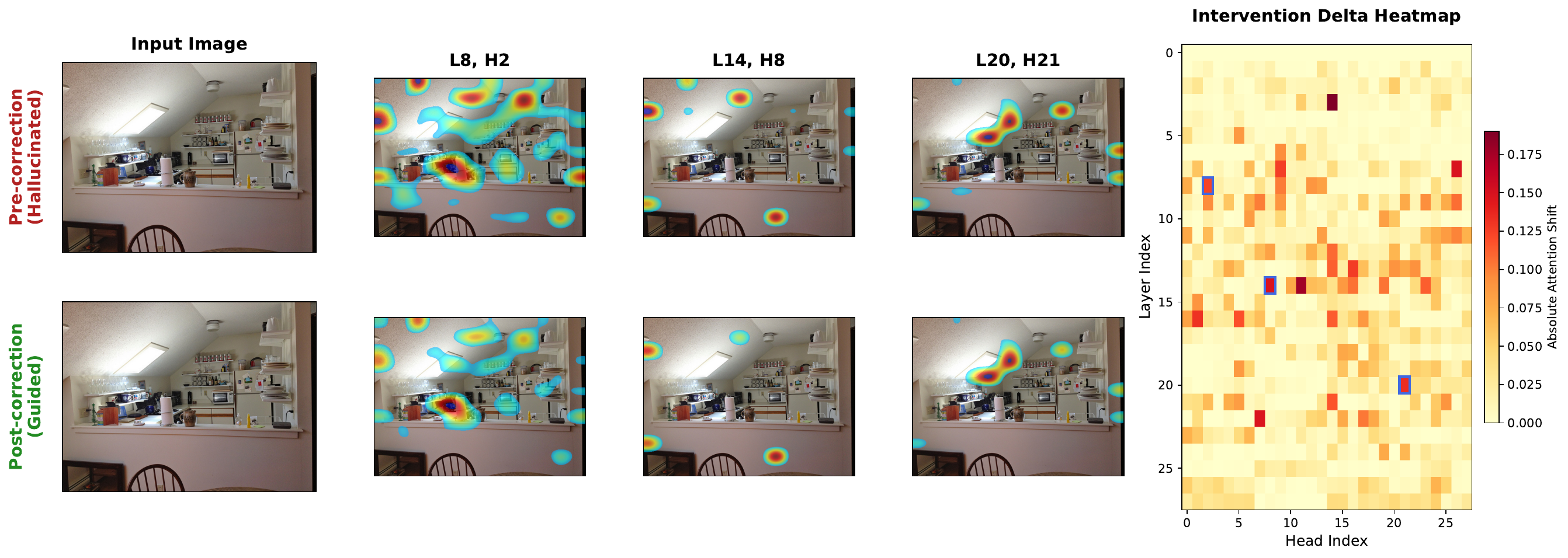}
\caption{Caption attention visualization (Qwen2.5-VL-7B). Top: pre-correction (hallucinated) attention maps at representative layer-head pairs. Bottom: post-correction (guided) attention maps. Right: intervention delta heatmap showing absolute attention shift per layer-head pair.}
\label{fig:supp_caption_attn}
\end{figure*}
 
\begin{figure*}[!htbp]
\centering
\includegraphics[width=0.93\textwidth]{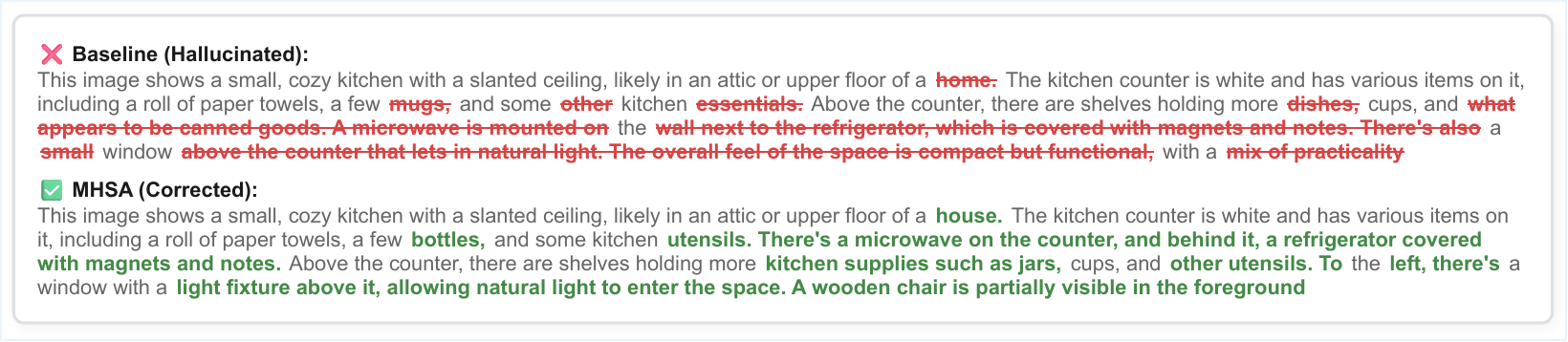}
\caption{Generated captions before and after MHSA correction (same image as \cref{fig:supp_caption_attn}). Red text indicates hallucinated content in the baseline output; green bold text indicates corrected or newly grounded content by MHSA.}
\label{fig:supp_caption_text}
\end{figure*}
 
\section{Additional Attention Visualization: Generative Task (Caption)}
\label{sec:appendix_attn_caption}
 
For generative tasks, we apply the token-level extension of MHSA. \cref{fig:supp_caption_attn} shows the per-token cross-modal attention at selected layer-head pairs before and after MHSA correction for a kitchen scene, together with the intervention delta heatmap across all layers and heads. The baseline model generates several hallucinated items (e.g., ``canned goods,'' ``mugs'') that are absent from the actual image. By monitoring the cross-modal attention at each autoregressive step, the DHCP-based discriminator identifies tokens with misaligned attention patterns. The MHSA generator then produces a sparse correction $\Delta\mathbf{A}$; the intervention delta heatmap reveals that these corrections primarily target specific intermediate layers and attention heads where hallucination-related misalignment is most pronounced. \cref{fig:supp_caption_text} presents the corresponding generated captions. The corrected caption removes non-existent objects and correctly identifies items (e.g., ``wooden chair'') that the baseline missed.

\end{document}